\documentclass{article}
\usepackage{iclr2027_conference,times}

\usepackage{amsmath,amssymb,mathtools}
\usepackage{array}
\usepackage{algorithm}
\usepackage{algpseudocode}
\usepackage{booktabs}
\usepackage{float}
\usepackage{graphicx}
\usepackage{microtype}
\usepackage{listings}
\usepackage{xcolor}
\usepackage{tikz}
\usetikzlibrary{arrows.meta,positioning,calc}
\usepackage{pgfplots}
\usepgfplotslibrary{groupplots}
\usepackage{tcolorbox}
\usepackage{hyperref}
\usepackage{xurl}

\definecolor{opineblue}{HTML}{0072B2}
\definecolor{opineorange}{HTML}{D98200}
\definecolor{opinegreen}{HTML}{008C72}
\definecolor{opinemagenta}{HTML}{CC79A7}
\definecolor{opinered}{HTML}{D55E00}
\definecolor{opineaction}{HTML}{6A3D9A}
\definecolor{opinelavender}{HTML}{e05e5e}
\pgfplotsset{
  compat=1.18,
  paper axis/.style={
    axis line style={black!65},
    tick style={black!65},
    tick label style={font=\scriptsize},
    label style={font=\scriptsize},
    title style={font=\small},
    grid style={black!10},
    legend style={font=\scriptsize,draw=none,fill=none}
  }
}

\hypersetup{
  colorlinks=true,
  linkcolor=blue!55!black,
  citecolor=blue!55!black,
  urlcolor=blue!55!black
}

\lstdefinestyle{conceptmap}{
  moredelim=**[is][\color{opineblue}]{<B>}{</B>},
  moredelim=**[is][\color{opineorange}]{<O>}{</O>},
  moredelim=**[is][\color{opinegreen}]{<G>}{</G>},
  moredelim=**[is][\color{opinemagenta}]{<M>}{</M>},
  moredelim=**[is][\color{opineaction}]{<A>}{</A>},
  moredelim=**[is][\color{opinelavender}]{<L>}{</L>}
}

\newcommand{\Obj}{\mathcal O}
\newcommand{\Types}{\mathcal T}
\newcommand{\SF}{\mathcal X}
\newcommand{\AF}{\mathcal A}
\newcommand{\IF}{\mathcal I}
\newcommand{\NF}{\mathcal N}

\newcommand{\Val}{\mathsf{Val}}
\newcommand{\Data}{\mathcal D}
\newcommand{\Effects}{\mathcal E}
\newcommand{\code}[1]{{\normalfont\ttfamily\detokenize{#1}}}

\newcommand{\sym}[1]{\mbox{\normalfont\ttfamily\detokenize{#1}}}
\newcounter{runningexample}
\newenvironment{runningexample}[2][]{%
  \refstepcounter{runningexample}%
  \begin{tcolorbox}[
    colback=black!3,
    colframe=black!28,
    colbacktitle=black!28,
    coltitle=white,
    title={\textbf{Running example: #2}},
    fonttitle=\small,
    fontupper=\small,
    boxrule=0.45pt,
    arc=2pt,
    left=6pt,
    right=6pt,
    top=5pt,
    bottom=5pt,
    #1
  ]
}{%
  \end{tcolorbox}
}
\newenvironment{runningexamplecontinued}[2][]{%
  \begin{tcolorbox}[
    colback=black!3,
    colframe=black!28,
    colbacktitle=black!28,
    coltitle=white,
    title={\textbf{Running example: #2}},
    fonttitle=\small,
    fontupper=\small,
    boxrule=0.45pt,
    arc=2pt,
    left=6pt,
    right=6pt,
    top=5pt,
    bottom=5pt,
    #1
  ]
}{%
  \end{tcolorbox}
}
\newcolumntype{L}[1]{>{\raggedright\arraybackslash}p{#1}}
\DeclareMathOperator{\Entropy}{H}
\DeclareMathOperator{\Dir}{Dir}

\DeclareMathOperator{\Pa}{Pa}

\newif\ifarxiv
\arxivtrue

\title{OPINE-World: Programmatic World Modeling with
Ontology-error-Prioritized Interactive Exploration for ARC-AGI-3}

\ifarxiv
\author{
David Courtis \quad Wenhao Li \quad Scott Sanner\\[2pt]
\normalfont
University of Toronto\\
Vector Institute for Artificial Intelligence\\
Toronto, Ontario, Canada\\[2pt]
\texttt{\{david.courtis,chriswenhao.li\}@mail.utoronto.ca}\\
\texttt{ssanner@mie.utoronto.ca}
}
  \iclrfinalcopy
\else
  \author{}
\fi

\begin{document}
\maketitle

\ifarxiv
  \lhead{Preprint}
\fi

\begin{abstract}

Learning a useful world model from minimal interaction is central to building agents that adapt to unfamiliar tasks. Programmatic world modeling approaches such as WorldCoder quickly learn transition models when given pre-supplied symbolic representations; however, such approaches are insufficient when the symbolic representation and goal are unknown and continually changing, and are intractable under large action spaces due to rigid planning heuristics. Extending the programmatic world modeling approach, OPINE-World additionally maintains an abstracted representation in which provisional objects, incomplete causal explanations, and unresolved interactions can exist in a Bayesian, exploration-centric hypothesis space. This enables learning an unknown, dynamically evolving state, transition, and goal. Exploration and program revision share this evolving abstraction, allowing new evidence to revise both the hypothesized mechanics (transition and reward behaviors) and their implementation. ARC-AGI-3 presents this open-world problem as a general-intelligence learning benchmark. Through our OPINE-World model learning paradigm, we improve the Relative Human Action Efficiency score of Claude Opus 4.8 high from 1.5\% to 78.4\% and GPT-5.5 high from 0.5\% to 84.3\% on the ARC-AGI-3 benchmark.

\end{abstract}

\section{Introduction}
\label{sec:intro}

\suppressfloats[t]
\begin{figure}[t]
\centering
\includegraphics[width=0.95\linewidth]{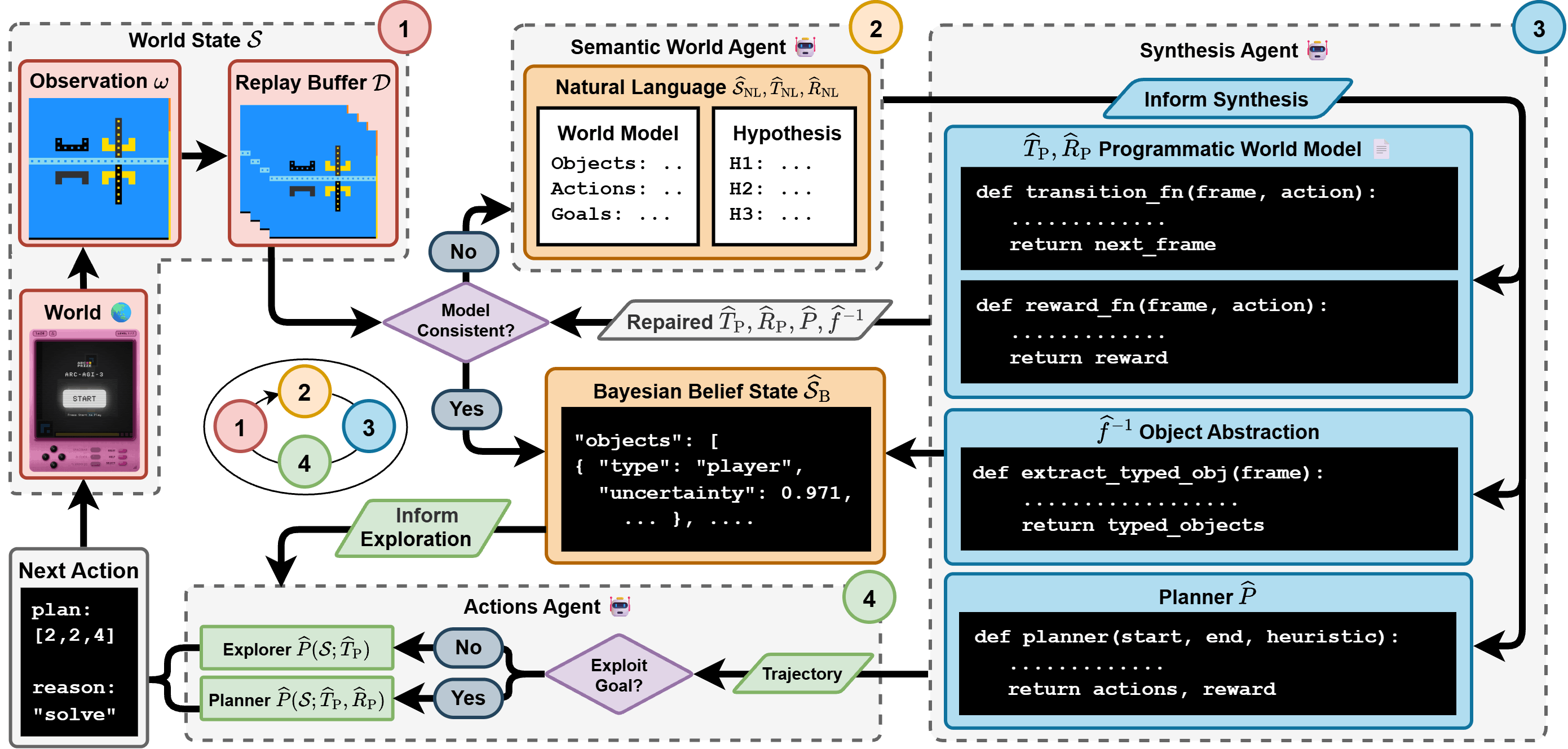}
\caption{\textbf{OPINE-World.} The world produces an observation \(\omega\), stored in replay buffer \(\Data\) and checked against the programmatic model. Model inconsistency repairs \((\widehat T_{\mathrm P},\widehat R_{\mathrm P},\widehat P,\widehat f^{-1})\). A consistent model routes a trajectory to the action agent, which uses the planner \(\widehat P(\mathcal S;\widehat T_{\mathrm P},\widehat R_{\mathrm P})\) to exploit, or the explorer \(\widehat P(\mathcal S;\widehat T_{\mathrm P})\) to resolve uncertainty with the NL and Bayesian belief state \(\widehat{\mathcal S}_{\mathrm B}\).}
\label{fig:harness}
\end{figure}


World model learning is essential for active reasoning and decision-making, but latent approaches rely on more training data than is feasible online \citep{sutton1991dyna,ha2018worldmodels,hafner2025dreamerv3}. Programmatic models such as WorldCoder learn executable transition and reward models over a pre-supplied symbolic representation \citep{tang2024worldcoder,poeworld2025} and reuse dynamics efficiently \citep{diuk2008oomdp,battaglia2018relational,kipf2020contrastive}, but rely on a rare assumption of accurate scene decomposition. Such approaches fail when the symbolic state and goal are simultaneously unknown and continually changing. Extending the world modeling approach of WorldCoder, our idea is to maintain two levels of representation: a qualitative, object-oriented natural language (NL) proto-representation of the world, and an executable world model in Python. In programmatic form, internal simulation and reward-directed planning to goal search become possible, whereas the NL proto-representation allows qualitative abstractions, enabling the management of hypothesis spaces and heuristics that require further exploration before conversion to programmatic form. Exploration and program revision share this evolving abstraction, allowing new evidence to revise both the hypothesized mechanics and their implementation in tandem at test time. The shared object abstraction additionally supports lifted Bayesian tracking of epistemic uncertainty over object--action effects. These beliefs identify sparsely tested or correlative effects, allowing for targeted exploration to learn causal relationships.

Using Claude Opus 4.8 high, OPINE-World clears 20 of 25 ARC-AGI-3 games, improving the Relative Human Action Efficiency (RHAE) score of the base model from 1.5\% to 78.40\%; using GPT-5.5 high, it clears 21 of 25 games, improving RHAE from 0.5\% to 84.27\%. Removal of the Bayesian epistemic layer \(\widehat{\mathcal S}_{\mathrm B}\) for exploration reduces RHAE by 20.24 points. We show that retaining our object-oriented proto-representation significantly reduces actions, synthesis rounds, and tokens during early online learning compared to a purely programmatic approach (Section ~\ref{sec:nl-results}), and enforces program structure that increases generalization and robustness (Section ~\ref{sec:structure-results}). In a direct comparison on WorldCoder compatible domains, OPINE-World produces action sequences closer to the optimal solution and solves more environments despite both systems having access to a planner (Section~\ref{sec:worldcoder-results}).

\paragraph{Contributions.}
\begin{itemize}
\item To avoid inefficient, low-diversity exploration patterns, we introduce an object-oriented Bayesian belief state \(\widehat{\mathcal S}_{\mathrm B}\) that converts uncertainty over type--action effects into targeted exploration priorities and proposes missing state distinctions.
\item To prevent early programmatic commitment and costly world model repairs, we show under ablations that \(\widehat{\mathcal S}_{\mathrm B}\)'s type-centric abstraction encourages more generalizable, maintainable, and repairable program patterns than monolithic, object-oriented, and naive methods.
\item To preserve non-executable hypotheses, we introduce a hybrid NL, Bayesian, and programmatic model state and show that the NL proto-representation improves initial online learning over a purely programmatic model.
\end{itemize}

By extending WorldCoder with qualitative and executable world models that improve through uncertainty-aware exploration, the OPINE-World learning loop increases world modeling capability far beyond the base model’s standalone performance and solves 80\% of public ARC-AGI-3 domains.


\section{Problem Setting: Open-world interaction}
\label{sec:problem-setting}
\label{sec:open-world-problem}

Consider an agent placed inside an unfamiliar environment. At step \(t\), the environment has a structured state \(s_t\in\mathcal S\), but only exposes a rendered abstraction \(\omega_t\). The agent selects an action \(a_t\) from a finite interface \(\mathcal A\), then receives a reward and the next observation \(\omega_{t+1}\):
\begin{equation}
  \omega_t=f(s_t),\qquad
  s_{t+1}=T(s_t,a_t),\qquad
  r_t=R(s_t,a_t,s_{t+1}).
  \label{eq:problem-mdp}
\end{equation}

\textbf{The structured state \(\mathcal S\), action semantics, transition function \(T\), reward function \(R\), and goal are all withheld.} In ARC-AGI-3, the agent can learn only from pixels, returned rewards, and its own interactions. Its evidence after \(t\) interactions is \(\Data_t=\{(\omega_\ell,a_\ell,r_\ell,\omega_{\ell+1})\}_{\ell<t}\). A later level may introduce a new object, relation, action, or reward condition; thereby, the objective is not to merely fit \(\Data_t\), but to efficiently acquire a model that continuously predicts and explores well enough to reach an unknown goal. This setting is \emph{open world} from the learner's perspective.
This framing tells us two sources of useful interaction. First, a reward-directed plan under optimism drives the agent to a state where its current model claims a useful consequence will occur. If the plan succeeds, the agent solves the task; if it fails, reality provides a goal-relevant counterexample. We call this \emph{goal-directed falsification}. Second, a information aware epistemic mechanism can explore an unresolved mechanic independent of reward paths. This is important for mechanics that can't appear in a reward-directed plan because the current model lacks the ontology needed to represent them. We call this \emph{ontology-error-prioritized interactive exploration}.

\section{Methodology}
\label{sec:method}

\subsection{Model representation}
\label{sec:system-overview}
\label{sec:target}
\begin{equation}
\widehat{\mathcal W}=\big(
  \underbrace{(\widehat{\mathcal S}_{\mathrm{NL}},\widehat T_{\mathrm{NL}},\widehat R_{\mathrm{NL}})}_{\text{NL model}},\;
  \underbrace{\widehat{\mathcal S}_{\mathrm B}}_{\text{Bayesian belief state}},\;
  \underbrace{(\widehat T_{\mathrm P},\widehat R_{\mathrm P},\widehat P,\widehat f^{-1})}_{\text{programmatic model}}\big).
  \label{eq:hybrid-world-model}
\end{equation}

OPINE-World represents the unknown state, transition, and reward through a unified NL, Bayesian, and executable Python system. At a high level, the representation has three parts:
\begin{enumerate}
\item \textbf{NL model \((\widehat{\mathcal S}_{\mathrm{NL}},\widehat T_{\mathrm{NL}},\widehat R_{\mathrm{NL}})\).} \(\widehat{\mathcal S}_{\mathrm{NL}}\) describes the ontology of object instances, inferred types, properties, and relations, \(\widehat T_{\mathrm{NL}}\) encodes qualitative and partially specified mechanics, and \(\widehat R_{\mathrm{NL}}\) encodes candidate goals and unresolved hypotheses (Section~\ref{sec:natural-language-world-model}).
\item \textbf{Bayesian belief state \(\widehat{\mathcal S}_{\mathrm B}\).} \(\widehat{\mathcal S}_{\mathrm B}\) quantifies object uncertainty from the observed effects of each object type, action, and context (Section~\ref{sec:eta}).
\item \textbf{Programmatic model \((\widehat T_{\mathrm P},\widehat R_{\mathrm P},\widehat P,\widehat f^{-1})\).} \(\widehat f^{-1}\) extracts object instances, inferred types, properties, and relations from \(\omega\) (Section~\ref{sec:observation-abstraction}). \(\widehat T_{\mathrm P}\) encodes a programmatic transition model and \(\widehat R_{\mathrm P}\) implements a programmatic reward model (Section~\ref{sec:programmatic-world-model}). The planner \(\widehat P\) rolls \((\widehat T_{\mathrm P},\widehat R_{\mathrm P})\) forward to pursue reward, or uses \(\widehat T_{\mathrm P}\) with \(\widehat{\mathcal S}_{\mathrm B}\) and the NL model for exploration (Section~\ref{sec:informative-interaction}).
\end{enumerate}

\subsection{Model learning loop}
\label{sec:model-learning-loop}

\begin{figure}[H]
\centering
\includegraphics[width=0.83\linewidth]{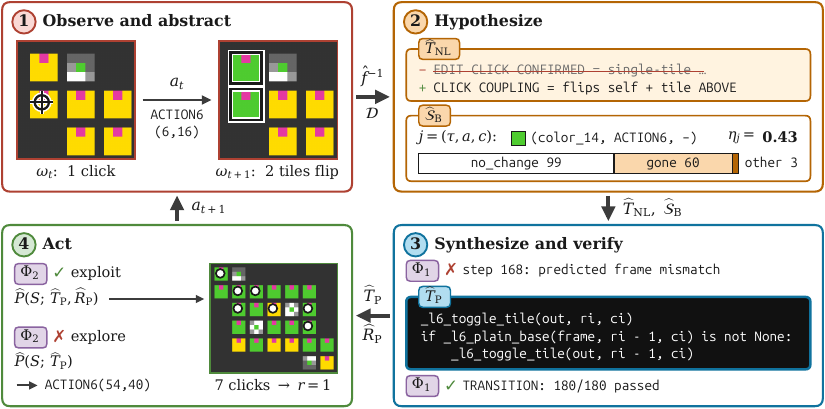}
\caption{\textbf{Example pass of the OPINE learning loop on \code{ft09}, }described in Appendix~\ref{app:ft09-example}.}
\label{fig:ft09-pipeline}
\end{figure}

The world learning loop performs the steps shown in Figure~\ref{fig:ft09-pipeline}:
\begin{enumerate}
\item \textbf{Observe and abstract.} Receive \((\omega_{t+1},r_t)\) after executing \(a_t\), append the transition to replay buffer \(\Data\), apply \(\widehat f^{-1}\), and associate objects across \(\omega_t\) and \(\omega_{t+1}\) (Section~\ref{sec:observation-abstraction}).
\item \textbf{Hypothesize.} Update \(\widehat{\mathcal S}_{\mathrm B}\) and revise the NL model (Section~\ref{sec:natural-language-world-model}).
\item \textbf{Synthesize and verify.} Compare the prediction with the observed truth. Treat disagreements as counterexamples and synthesize a revised \((\widehat T_{\mathrm P},\widehat R_{\mathrm P},\widehat P,\widehat f^{-1})\) under new observations (Section~\ref{sec:cegis}).
\item \textbf{Act.} For a consistent model, use \(\widehat P(\mathcal S;\widehat T_{\mathrm P},\widehat R_{\mathrm P})\) to pursue a goal or \(\widehat P(\mathcal S;\widehat T_{\mathrm P})\), informed by the NL model and \(\widehat{\mathcal S}_{\mathrm B}\), to explore, and execute the next actions, producing a new observation \(\omega\) (Section~\ref{sec:informative-interaction}).
\end{enumerate}

\subsubsection{Step 1: Observe and abstract}
\label{sec:observation-abstraction}
\label{sec:state-extractor}

At each step, \(\widehat f^{-1}\) turns the rendered observation \(\omega_t\) into the object-oriented state:
\begin{equation}
\widehat s_t=\widehat f^{-1}(\widehat s_{t-1},\omega_t).
\label{eq:object-state-main}
\end{equation}
The resulting object records are the common observable state read by \(\widehat T_{\mathrm P}\), \(\widehat R_{\mathrm P}\), the planner, the verifier, and the Bayesian belief state \(\widehat{\mathcal S}_{\mathrm B}\). This state reconstructs the objects, their inferred types and properties, and relations from the current observation and history. (Appendix~\ref{app:rddl-state}).

\subsubsection{Step 2: Hypothesize}
\label{sec:natural-language-world-model}

The NL model \((\widehat{\mathcal S}_{\mathrm{NL}},\widehat T_{\mathrm{NL}},\widehat R_{\mathrm{NL}})\) acts as a proto-representation of our MDP. It can represent partial claims such as \emph{``colour x may be an important confounding variable for object y''}. These uncertainties are quantitatively expressed in a Bayesian belief state \(\widehat{\mathcal S}_{\mathrm B}\), which records empirical effect distributions for object type, action, and observable context from a deterministic, replay-based extractor. A mixed distribution of such observations in a deterministic environment indicates that the current model fails to distinguish a confounding variable. \(\Data\) and \(\widehat{\mathcal S}_{\mathrm B}\) update on every interaction. The action agent revises \((\widehat{\mathcal S}_{\mathrm{NL}},\widehat T_{\mathrm{NL}},\widehat R_{\mathrm{NL}})\) when \(\omega\) provides useful or surprising information. Appendix~\ref{app:ft09-example} explains the \code{ft09} game used in the examples.

\begin{runningexample}[top=2pt]{State in \(\widehat{\mathcal S}_{\mathrm{NL}}\) and \(\widehat{\mathcal S}_{\mathrm B}\) form}
\label{ex:ft09-state-parallel}
\noindent
\begin{minipage}[t]{0.49\linewidth}
\vspace{0pt}
\centering
\textbf{\(\widehat{\mathcal S}_{\mathrm{NL}}\): qualitative}\par
\begin{lstlisting}[
  style=conceptmap,
  basicstyle=\ttfamily\fontsize{5.8}{6.4}\selectfont,
  aboveskip=5pt,
  belowskip=0pt,
  framesep=1.5pt
]
<B>TYPE</B>
<B>c1 is an answer_cell.</B>
<O>PROPERTY</O>
<O>color(c1) is usually yellow, but may be blue rarely?</O>
<G>RELATION</G>
<G>c2 is a pictured, occupied  neighbor of c1.</G>
<M>DERIVED STATE</M>
<M>clicking c1 usually changes one other answer cell.</M>
\end{lstlisting}
\end{minipage}\hfill
\begin{minipage}[t]{0.49\linewidth}
\vspace{0pt}
\centering
\textbf{\(\widehat{\mathcal S}_{\mathrm B}\): quantitative}\par
\begin{lstlisting}[
  style=conceptmap,
  basicstyle=\ttfamily\fontsize{5.8}{6.4}\selectfont,
  aboveskip=5pt,
  belowskip=0pt,
  framesep=1.5pt
]
{
<B>  "object": {"id": "c1", "type": {"answer_cell": 1}},</B>
<O>  "property": {"color(c1)": {"y": 0.98, "b": 0.02}},</O>
<G>  "relation": { "brush_neighbor(c1,c2)": 0.82, </G>
<G>     "no_neighbour": 0.18 },</G>
<M>  "derived": {</M>
<M>    "cell_brush_arity(c1)": 0.75}</M>
}
\end{lstlisting}
\end{minipage}

\smallskip
The probabilities schematically illustrate the current belief given \(\Data\). Appendix~\ref{app:eta} explains the lifted attribution to effects and contexts.
\end{runningexample}

\paragraph{Bayesian effect uncertainty and state refinement.}
\label{sec:eta}
\(\widehat{\mathcal S}_{\mathrm B}\) groups object transitions into their observed effect under unique value combinations of \(j=(\tau,a,c)\), where \(\tau\) is an inferred object type, \(a\) is the action, and \(c\) summarizes relevant context. Each group keeps a Laplace-smoothed distribution over its observed effects, scored by its normalized entropy \(\eta_j\) (Appendix~\ref{app:eta}). We use the posterior predictive entropy \(\eta_j\) as a diagnostic of sparse evidence or confounds under the deterministic modeling assumption. To target confounds, the groups with the largest total unexplained entropy (\(\eta_j\) weighted by the group's observation count) are passed to synthesis where a \(\eta_j\) minimizing split is kept if discovered. Each mixed group is then divided by the values of its resulting contexts, and each subgroup is scored as \(1-\frac{\text{count of its most common outcome}}{1+\text{number of observations}}\) and provided as an actor-facing exploration priority (Appendix~\ref{app:eta}).

\subsubsection{Step 3: Synthesize and verify}
\label{sec:programmatic-world-model}

\paragraph{Programmatic transition model.}
\label{sec:programmatic-transition-model}
The transition program translates the qualitative mechanics \(\widehat T_{\mathrm{NL}}\) into an executable successor:
\begin{equation}
\widehat T_{\mathrm P}:
  \widehat{\mathcal S}\times\mathcal A\rightarrow\widehat{\mathcal S},
\qquad
\widehat s'=\widehat T_{\mathrm P}(\widehat s,a).
\label{eq:program-transition}
\end{equation}
The learned update for each fluent mutable state may read object types, current attributes, grounded action arguments, and derived \code{@fluent} relations -- these fluents are synthesized and mechanically attached to objects to determine the relevant context for Bayesian epistemic uncertainty tracking. This interface therefore contains action grounding, the conditions under which an effect applies, and the resulting changes. For our factored-dynamics interpretation refer to Appendix~\ref{app:cpfs}.

\(\widehat T_{\mathrm{NL}}\) may retain several plausible or partially specified mechanics while \(\widehat T_{\mathrm P}\) is deterministic and executable. Repeated application of \(\widehat T_{\mathrm P}\) supports planning.

\begin{runningexamplecontinued}[top=2pt]{Transitions}
\label{ex:ft09-transition-parallel}
\vspace*{-3pt}
\noindent
\begin{minipage}[t]{0.52\linewidth}
\vspace{0pt}
\centering
\textbf{\(\widehat T_{\mathrm{NL}}\): qualitative}\par
\begin{lstlisting}[
  style=conceptmap,
  basicstyle=\ttfamily\fontsize{5.8}{6.4}\selectfont,
  aboveskip=5pt,
  belowskip=0pt,
  framesep=1.5pt
]
<A>ACTION6(x,y) is a click.</A>

<B>Acts only when (x,y) hits an answer cell c.</B>

<G>The marks on c denote neighboring answer cells.</G>

<O>An object's arity is the number of affected neighbors.</O>

<M>Advance c and those neighbors one palette step. Keep</M>
<M>the rest.</M>

<L>Hypothesis: The palette wraps around when exhausted.</L>
\end{lstlisting}
\end{minipage}\hfill
\begin{minipage}[t]{0.46\linewidth}
\vspace{0pt}
\centering
\textbf{\(\widehat T_{\mathrm P}\): quantitative}\par
\begin{lstlisting}[
  style=conceptmap,
  basicstyle=\ttfamily\fontsize{5.8}{6.4}\selectfont,
  aboveskip=5pt,
  belowskip=0pt,
  framesep=1.5pt
]
@fluent
<O>def cell_brush_arity(c, state):</O>
<O>    return len(affected_neighbors(c, state))</O>
def apply_click(state, action):
<M>    nxt = deepcopy(state)</M>
<A>    if action["action_id"] != 6: return nxt</A>
<B>    hit = hit_answer_cell(nxt, action)</B>
<B>    if hit is None: return nxt</B>
<G>    nbrs = affected_neighbors(hit, nxt)</G>
<M>    for c in [hit, *nbrs]:</M>
<M>        advance_palette(c, nxt)</M>
    return nxt
\end{lstlisting}
\end{minipage}
\end{runningexamplecontinued}

\paragraph{Programmatic reward model.}
\label{sec:programmatic-reward-model}
\label{sec:grounding}
For sparse completion rewards, the program grounds the current goal hypothesis as a boolean predicate \(\widehat G\) and uses:
\begin{equation}
\widehat R_{\mathrm P}(\widehat s,a,\widehat s')
=r^{0}+\big(r^{+}-r^{0}\big)
 \widehat G(\widehat s,a,\widehat s'),
\qquad \widehat G\in\{0,1\}.
\label{eq:program-reward}
\end{equation}
The reward program \(\widehat R_{\mathrm P}:\widehat{\mathcal S}\times\mathcal A\times\widehat{\mathcal S}\rightarrow\mathbb R\) assigns a prediction to an transition dependent on a precondition $\widehat s$, an action $a$ and the resulting state $\widehat s'$.
After a reward observation or a revised qualitative goal, a goal-grounding agent converts \(\widehat R_{\mathrm{NL}}\) into a bounded set of Boolean requirements over extracted objects given a new level into \(\widehat R_{\mathrm P}\). Appendix~\ref{app:ft09-reward} illustrates the corresponding NL and Python representations for \code{ft09}.

\paragraph{Predictive verification \texorpdfstring{\(\Phi_1\)}{Phi 1}.}
\label{sec:cegis}
Let \(\widehat s_\ell\) be the extracted state at step \(\ell\). The current executable hypothesis is replay-consistent when
\begin{equation}
\Phi_1(\widehat f^{-1},\widehat T_{\mathrm P},\widehat R_{\mathrm P};\Data_t)\Longleftrightarrow
\bigwedge_{\ell<t}\!\left[
\widehat T_{\mathrm P}(\widehat s_\ell,a_\ell)=\widehat s_{\ell+1}
\;\land\;
\widehat R_{\mathrm P}(\widehat s_\ell,a_\ell,\widehat s_{\ell+1})=r_\ell
\right].
\label{eq:replay}
\end{equation}
Passing \(\Phi_1\) means that the selected fields and rewards are consistent with observed transitions (Appendix~\ref{app:verification-details}). Each disagreement is treated as a counterexample for the next candidate program, and counterexample-guided inductive synthesis (CEGIS) synthesizes a revised programmatic model under sufficient batches \citep{solarlezama2006sketching,jha2010oracle,alur2013sygus} and reconciles the NL model.

\subsubsection{Step 4: Act}
\label{sec:informative-interaction}

The planner \(\widehat P\) searches the programmatic model for a sequence \(\pi=(a_0,\ldots,a_{L-1})\).  Starting from \(s^0=\widehat s_t\), it rolls candidate actions through \(\widehat T_{\mathrm P}\) as \(s^{k+1}=\widehat T_{\mathrm P}(s^k,a_k)\). Define
\begin{equation}
\Phi_2\!\left(
  \pi;s^0,\widehat{T}_{\mathrm{P}},\widehat{R}_{\mathrm{P}}
\right)
\Longleftrightarrow
\max_{0\le k<L}
\widehat{R}_{\mathrm{P}}(s^k,a_k,s^{k+1})>0.
\label{eq:reward-directed-plan}
\end{equation}
Satisfying \(\Phi_2\) gives a \emph{reward-directed plan}. There are two informative outcomes. 1) The predictions hold and the plan reaches reward. 2) A prediction fails and \(\Phi_1\) returns a task-relevant counterexample for model repair. Thus, this is an active model-testing mechanism. Before and during use, predicted successors are checked by \(\Phi_1\) and subjected to CEGIS. A prediction mismatch stops the current plan and opens a short collection window for related counterexamples (Appendix~\ref{app:experiments}).

An \emph{epistemic experiment} is an additional source of experience. Posterior uncertainty in \(\widehat{\mathcal S}_{\mathrm B}\) identifies under-sampled or conflicting object--action effects, and the semantic world agent uses \((\widehat{\mathcal S}_{\mathrm{NL}},\widehat T_{\mathrm{NL}},\widehat R_{\mathrm{NL}})\) to propose actions that aim to find the confounder. This experiment is useful when no \(\Phi_2\)-satisfying plan exists.

\section{Experimental Setup}
\label{sec:experiments}


\paragraph{RQ1: Does our Bayesian epistemic guidance \(\widehat{\mathcal S}_{\mathrm B}\) improve exploration?}
\label{rq:online-experiment-selection}
Alongside a comparison against other systems, we compare the full system against an \(\widehat{\mathcal S}_{\mathrm B}\) ablation, and an $\epsilon$-greedy baseline (Appendix~\ref{app:epsilon-400}) on the 25 public ARC-AGI-3 games with a 2000-action budget per game, or on a stall of more than 500 actions per level. We measure the per-game difference in RHAE (Appendix~\ref{app:scoring}) and inspect the action reasoning traces to identify the failure sources (Appendix~\ref{app:exploration-failures}).

\paragraph{RQ2: How does the program structure used to encode transition semantics affect generalizability, adaptability, and maintainability (code metrics) of the programmatic world model?}
\label{rq:predictive-programs}
Under a fixed action trajectory for all runs, we both prompt-restrict and code-structure-restrict the programmatic world model outputted by the synthesizer agent in various ways: 

\begin{enumerate}
\item \textbf{Monolithic Code Structure.} Classic MDP framing, where the whole world's transitions are described in a single monolithic transition function.
\item \textbf{Type-Oriented Code Structure.} An object-oriented framing where transition functions for an object type are defined within each object class, or a type-relational class.
\item \textbf{\(\widehat{\mathcal S}_{\mathrm B}\) Code Structure (Unablated).} A hybrid framing encouraged naturally by the Bayesian Belief State where transition functions for an object type are encouraged to be defined within each object type's class only where easily definable through the Bayesian Belief State.
\item \textbf{Unrestricted (No \(\widehat{\mathcal S}_{\mathrm B}\)).} A completely unconstrained code synthesis for control purposes.
\end{enumerate}
After confirming that the interventions produce different program structures (Appendix~\ref{app:structure-metrics}), we measure generalizability, adaptability, and maintainability. Adaptability is assessed through locality and synthesis cost of code revisions from a modified environment, measured by changed source lines, engine rewrite percentage, functions changed, synthesis rounds, and output tokens. Generalizability is measured as the prediction accuracy on unseen state transitions across these modified environments. To facilitate these measurements, we change an existing object mechanic (M1), introduce new objects and relations (M2), and change a global rule (M3) (Appendix~\ref{app:repair-controls}).


\paragraph{RQ3: Will ablating the NL world model \((\widehat{\mathcal S}_{\mathrm{NL}},\widehat T_{\mathrm{NL}},\widehat R_{\mathrm{NL}})\) and using a pure code representation perform as well as our hybrid NL and programmatic world model approach?}
\label{rq:natural-language-model}
We compare against an NL ablation that prevents the generation of an explicit NL world model as files. Because the synthesis agent's context is reset at each stage, we can consider this as removing the persistent NL representation of the world. We hypothesize that NL can better represent hypotheses and unresolved correlative observations, which should increase the RHAE and decrease the model-revision cost. We run on (\code{cd82}, \code{cn04}, \code{ft09}, \code{g50t}, \code{tn36}) with a 2,000-action budget, and an early stopping at 5\% RHAE.

We use \textbf{Claude Opus 4.8 high 1m} and \textbf{GPT-5.5 high} under Docker isolation from game source and network access. RQ2 and RQ3 studies use Claude Opus 4.8 low; RQ2 uses \code{g50t} as a demanding case study combining history-dependent dynamics, interacting object types, and progressively introduced mechanics (Appendix~\ref{app:g50t-motivation}). Both systems in the WorldCoder comparison use a weaker GPT-6 Luna so the domains don’t saturate. The supplementary material contains the full implementation of OPINE-World (Appendix~\ref{app:experiments}), and ablation details (Appendix~\ref{app:ablation-scope}).
\section{Results}
\label{sec:results}
\subsection{Overall benchmark performance}
\label{sec:overall-results}

\begin{table}[H]
\caption{\textbf{RHAE of OPINE-World, and other systems.} Per-game results are in Appendix~\ref{app:benchmark-results}.}
\label{tab:external}
\centering
\scriptsize
\begin{tabular}{@{}lrrrrrrr@{}}
\toprule
& \multicolumn{3}{c}{OPINE-World (effort high)} & \multicolumn{4}{c}{Other systems (GPT-5.5 high)} \\
\cmidrule(lr){2-4}\cmidrule(l){5-8}
Metric & \shortstack{GPT-5.5} & \shortstack{Opus 4.8} & \shortstack{No \(\widehat{\mathcal S}_{\mathrm B}\) (Opus 4.8)} & baseline1 & DreamTeam & NOOA & \shortstack{Continual} \\
\midrule
Mean \(\pm\) SE & \textbf{84.3 \(\pm\) 3.5} & 78.4 \(\pm\) 6.3 & 58.2 \(\pm\) 8.3 & 63.7 \(\pm\) 7.4 & 38.1 \(\pm\) 7.6 & 50.2$^{\dagger}$ & 22.2 \(\pm\) 5.9 \\
Levels completed & \textbf{176/183} & 160/183 & 126/183 & 146/183 & 98/183 & 118/183 & 86/183 \\
Games completed & \textbf{21/25} & 20/25 & 12/25 & 14/25 & 6/25 & -- & 5/25 \\
\bottomrule
\end{tabular}
\end{table}

OPINE-World achieves the highest RHAE among comparators and completes the most games (Table~\ref{tab:external}, per-game results in Appendix~\ref{app:benchmark-results}). On GPT-5.5 high, OPINE-World also costs less than baseline1 and scores higher at every wall-clock budget (Appendix~\ref{app:context-accounting}). On games completed by both OPINE-World and the strongest comparator, the main advantage is level completion rather than efficiency, suggesting that broader discovery of mechanics contributes more to the aggregate gain than shorter solutions to familiar tasks (RQ1).

\label{rq:worldcoder}
We additionally run OPINE-World and WorldCoder \citep{tang2024worldcoder} on WorldCoder's MiniGrid and Sokoban domains, and give each agent information exposed in the original WorldCoder benchmark, with each run capped at 24 hours of planner CPU compute time on a Ryzen 7600X. We measure actions relative to the optimal plan and episodes to mastery.

We compare against non-ensemble, structured approaches to reasoning that allow a scientific comparison, analysis of design choices, and their contribution to LLM performance. Therefore, we omit the comparison of methods that were built on the structure of the OPINE-World, pure black-box approaches, and methods that require a foundation model with a significant difference in base capability, as these contributions are beyond our ability to explain.

\label{sec:worldcoder-results}

\begin{table}[h]
\caption{\textbf{WorldCoder Comparisons (GPT-6 Luna low).} Actions are relative to the optimal plan. -- indicates failure due to planner timeout on a 24h search budget.}
\label{tab:minigrid}
\centering
\scriptsize
\setlength{\tabcolsep}{4pt}
\begin{tabular}{@{}lrrcrrcrr@{}}
\toprule
& \multicolumn{2}{c}{\shortstack{First episode (\(\times\) optimal)}} && \multicolumn{2}{c}{\shortstack{Solved episodes (\(\times\) optimal)}} && \multicolumn{2}{c}{\shortstack{Episodes to mastery}} \\
\cmidrule(lr){2-3}\cmidrule(lr){5-6}\cmidrule(l){8-9}
Environment & WorldCoder & OPINE-World && WorldCoder & OPINE-World && WorldCoder & OPINE-World \\
\midrule
\multicolumn{9}{@{}l}{\textit{MiniGrid}} \\
Empty-5x5 & 3.33 \(\pm\) 0.27 & \textbf{1.67 \(\pm\) 0.29} && \textbf{1.45 \(\pm\) 0.02} & 1.52 \(\pm\) 0.35 && 11.0 & 11.0 \\
DoorKey-5x5 & 2.45 \(\pm\) 0.28 & \textbf{1.06 \(\pm\) 0.06} && 1.27 \(\pm\) 0.12 & \textbf{1.08 \(\pm\) 0.04} && \textbf{11.0} & 13.3 \(\pm\) 2.3 \\
Unlock & 2.13 \(\pm\) 0.15 & \textbf{1.04 \(\pm\) 0.04} && 1.32 \(\pm\) 0.05 & \textbf{1.10 \(\pm\) 0.04} && 11.0 & 11.0 \\
Fetch-6x6-N2 & 3.93 \(\pm\) 0.77 & \textbf{1.07 \(\pm\) 0.07} && 1.59 \(\pm\) 0.04 & \textbf{1.21 \(\pm\) 0.14} && 11.0 & 11.0 \\
UnlockPickup & -- & \textbf{1.25} && -- & \textbf{1.31 \(\pm\) 0.10} && -- & \textbf{11.0} \\
BlockedUnlockPickup & -- & \textbf{1.53 \(\pm\) 0.12} && -- & \textbf{1.60 \(\pm\) 0.19} && -- & \textbf{15.0 \(\pm\) 4.0} \\
\midrule
\multicolumn{9}{@{}l}{\textit{Sokoban}} \\
Small, 2 boxes & 7.83 \(\pm\) 0.83 & \textbf{3.11 \(\pm\) 0.44} && 1.41 \(\pm\) 0.20 & \textbf{1.23 \(\pm\) 0.01} && \textbf{12.0 \(\pm\) 1.0} & 16.7 \(\pm\) 3.0 \\
Small, 3 boxes & 1.05 \(\pm\) 0.05 & \textbf{1.00 \(\pm\) 0.00} && 1.04 \(\pm\) 0.04 & 1.04 \(\pm\) 0.02 && -- & \textbf{15.7 \(\pm\) 4.2} \\
\bottomrule
\end{tabular}
\end{table}

In a direct comparison on WorldCoder-compatible domains, OPINE-World produces action sequences closer to the optimal solution despite both systems having access to a planner (Table~\ref{tab:minigrid}). WorldCoder's planner fails to find a reward-reaching plan in UnlockPickup due to the absence of synthesized heuristics and cannot progress to BlockedUnlockPickup, while OPINE-World eventually masters all six environments. On Sokoban, WorldCoder masters the 2-box environment in fewer episodes; however, even with a gridsearch across a 1000--12000 node depth limit for its MCTS planner, its most successful configuration exhausts its generous 24h search budget on the 3-box layouts and is unable to progress. This suggests that OPINE-World provides more efficient and computable solutions to simpler domains (Appendix~\ref{app:worldcoder-limits}).

\subsection{RQ1: Effect of Bayesian guidance on online discovery}
\label{sec:exploration-results}

\begin{figure}[h]
\centering
\includegraphics[width=0.82\linewidth]{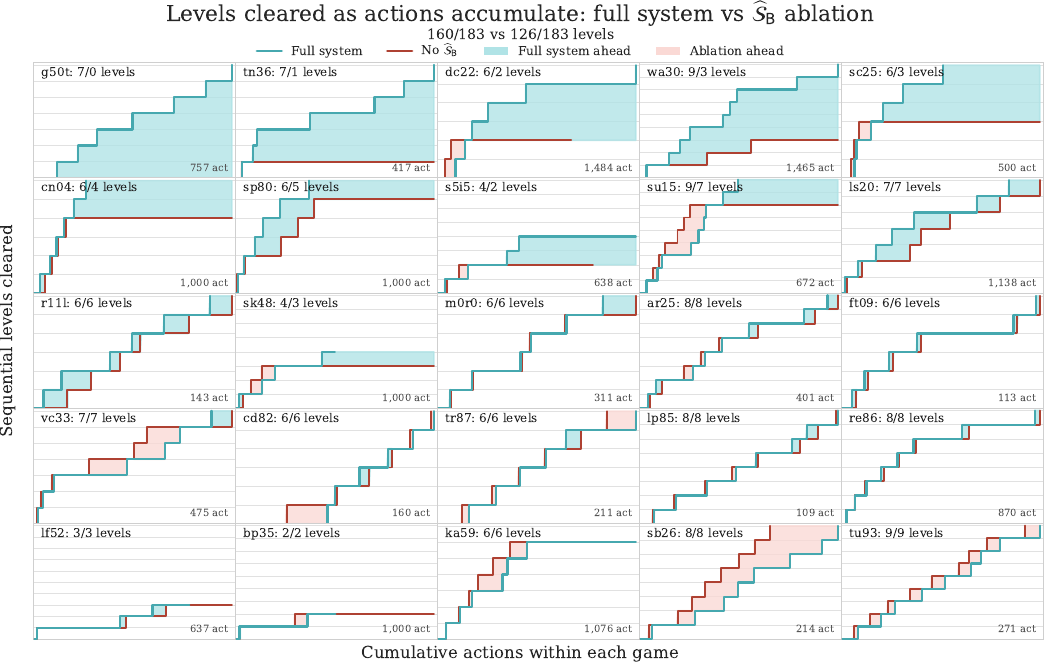}
\caption{\textbf{Online progression with and without Bayesian guidance.} Each step marks a completed level. Blue or red shading marks intervals where the denoted system leads. Ordered by blue area.}
\label{fig:rq1-progression}
\end{figure}

Removal of \(\widehat{\mathcal S}_{\mathrm B}\) reduces RHAE from 78.4 \(\pm\) 6.3 to 58.2 \(\pm\) 8.3, and games completed from 20 to 12 (Claude Opus 4.8 high). The largest gains over the \(\widehat{\mathcal S}_{\mathrm B}\) ablation occur when the agent fails to discover a mechanic needed for progress (Figure~\ref{fig:rq1-progression}). In \code{g50t} and \code{dc22}, the ablated agent repeats similar sequences or configurations while unresolved interactions remain untested. In \code{cn04}, it fails to understand confounds around button mechanics, and in \code{sc25}, it misses an earlier projectile mechanic. Missing an interaction required by the solution prevents level completion, even if the agent has learned the other mechanics correctly, indicating that such exploratory oversights can be catastrophic. These failures suggest that Bayesian guidance helps the agent move beyond familiar actions when its current model fails to explain how to make progress.

The effect can extend beyond the level in which a mechanic is first encountered. In \code{ls20}, the ablated agent fails to transfer a tested object-class property; in \code{sp80}, insufficient early probing leaves an incorrect mechanic in the model used for later planning. In some games, additional exploration is not useful, especially when exploration targets uncertainties that are entirely irrelevant to the goal; however, this often cannot be known in advance. 

We additionally measure against a naive exploration method: an $\epsilon$-greedy exploration baseline (Figure~\ref{fig:rq1-epsilon-400}). We see that random object exploration is insufficient to provide valuable evidence, and often is detrimental to goal planning if sufficient knowledge has been accrued. This motivates the case for a more principled exploration method such as \(\widehat{\mathcal S}_{\mathrm B}\). Detailed case-by-case review and trace excerpts are provided in Appendices~\ref{app:exploration-failures} and~\ref{app:hypothesis-traces}.


\begin{table}[H]
\caption{\textbf{\(\widehat{\mathcal S}_{\mathrm B}\) encourages stronger exploration.}}
\label{tab:rq1-acquisition}
\centering
\scriptsize
\setlength{\tabcolsep}{4pt}
\begin{tabular}{@{}lrrcrrcrrcrrcrr@{}}
\toprule
& \multicolumn{2}{c}{\shortstack{Levels\\completed}} && \multicolumn{2}{c}{\shortstack{Unique effects\\discovered}} && \multicolumn{2}{c}{\shortstack{Discovered effects\\modeled (\%)}} && \multicolumn{2}{c}{\shortstack{Repeated action\\sequences (\%)}} \\
\cmidrule(lr){2-3}\cmidrule(lr){5-6}\cmidrule(lr){8-9}\cmidrule(lr){11-12}
Game & Full & \(\widehat{\mathcal S}_{\mathrm B}\) Abl. && Full & \(\widehat{\mathcal S}_{\mathrm B}\) Abl. && Full & \(\widehat{\mathcal S}_{\mathrm B}\) Abl. && Full & \(\widehat{\mathcal S}_{\mathrm B}\) Abl. \\
\midrule
\code{cn04} & 6 & 4 && 101 & 60 && 100 & 100 && 1.1 & 6.1 \\
\code{dc22} & 6 & 2 && 203 & 50 && 100 & 100 && 9.4 & 10.3 \\
\code{g50t} & 7 & 0 && 44 & 28 && 100.0 & 75.0 && 15.2 & 32.8 \\
\code{s5i5} & 4 & 2 && 80 & 47 && 100 & 100 && 6.0 & 21.5  \\
\code{sc25} & 6 & 3 && 70 & 22 && 100 & 100 && 5.9 & 7.9 \\
\code{sp80} & 6 & 5 && 58 & 49 && 100.0 & 100.0 && 2.2 & 4.6 \\
\code{su15} & 9 & 7 && 93 & 62 && 100 & 100 && 3.3 & 17.9 \\
\code{tn36} & 7 & 1 && 36 & 11 && 100.0 & 100.0 && 7.7 & 17.0 \\
\code{wa30} & 9 & 3 && 235 & 120 && 98.7 & 92.4 && 3.3 & 11.5  \\
\midrule
Total & \textbf{60} & 27 && \textbf{920} & 449 && \textbf{99.9} & 96.4 && \textbf{6.0} & 14.4 \\
\bottomrule
\end{tabular}
\end{table}

While RHAE measures task success under a fixed action budget, making it a suitable measure of action efficiency, to look at exploration more directly, we also report the share of repeated action sequences in the nine games where the systems differ in level performance (Table~\ref{tab:rq1-acquisition}). A high share indicates a lack of progress or a narrow trajectory. While both systems model nearly all effects they observe (99.9\% and 96.4\%), the ablation observes fewer effects (449 vs. 920) and repeats more action sequences (14.4\% vs. 6.0\%). The ablation's losses therefore come from weaker acquisition of evidence rather than modeling failure.

\subsection{RQ2: Effect of program structure on prediction and revision}
\label{sec:structure-results}

\begin{figure}[t]
\centering
\includegraphics[width=0.82\linewidth]{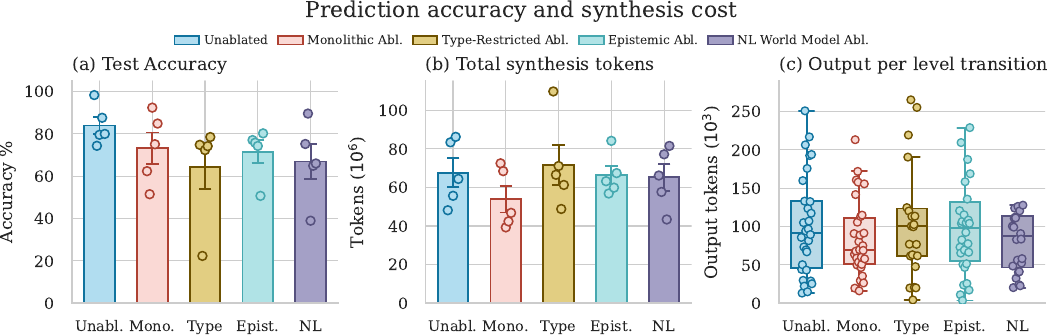}
\caption{\textbf{Prediction and synthesis cost under same evidence.} Error bars show standard errors.}
\label{fig:rq2-fixed}
\end{figure}

With the same evidence supplied to synthesis, \(\widehat{\mathcal S}_{\mathrm B}\) Code Structure (Unablated) has the highest accuracy on unseen transitions across the RQ2 conditions (Figure~\ref{fig:rq2-fixed}). The Type-Oriented variant performs worst despite using more tokens on average. Upon inspection, we found that successful models encoded object transitions in a shared monolithic structure because heavily coupled effects are difficult to assign to individual objects. As a result, programs that were learned with \(\widehat{\mathcal S}_{\mathrm B}\) Code Structure retained object-indexed state while allowing interactions to be resolved jointly, suggesting that the organization of transition rules should remain flexible even when the state is best represented as object-centric.

\begin{figure}[h]
\centering
\includegraphics[width=0.82\linewidth]{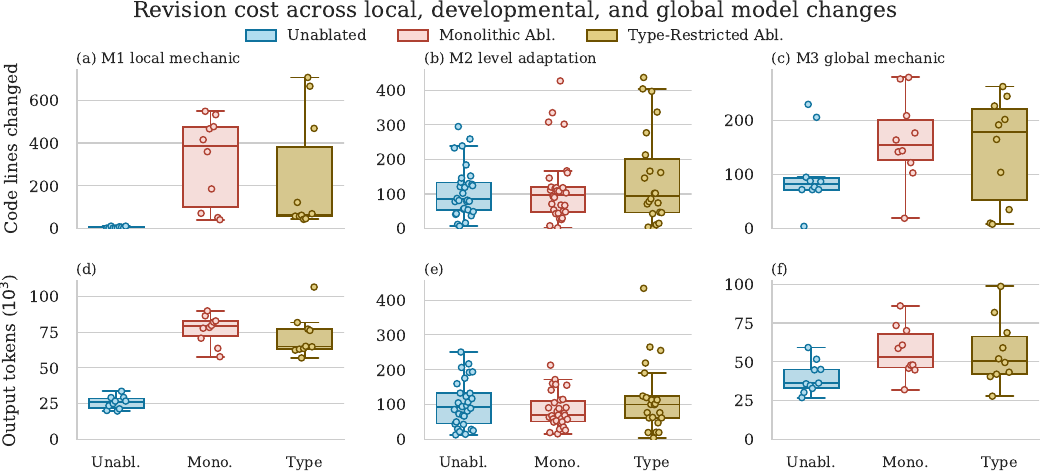}
\caption{\textbf{Revision costs for local changes, new level mechanics, and global changes.}}
\label{fig:rq2-revisions}
\end{figure}

The clearest revision advantage for \(\widehat{\mathcal S}_{\mathrm B}\) Code Structure occurs when a local mechanic changes (M1): it edits fewer definitions and requires less rewriting, fewer synthesis rounds, and fewer output tokens than Monolithic Code Structure and Type-Oriented Code Structure (Figure~\ref{fig:rq2-revisions}). Interestingly we see that the type-oriented structure has a large variance. Upon inspection, these costly outliers were due to overcommitment to structure, resulting in wrongly abstracted typings during periods of uncertainty. These are very expensive to repair and were avoided with the type-uncertainty information from \(\widehat{\mathcal S}_{\mathrm B}\).

However, when later levels introduce new mechanics (M2), we see that a new level usually requires significant changes even across multiple types, reducing the advantage of type abstraction. Interestingly, we found that the monolithic system has less structure, giving it an innate advantage for organizational cost that was counteracted by the code's difficulty to maintain. Although costs are similar across methods, the superior generalization of the unablated system shows that such adaptation is more robustly incorporated. Additionally, we see in M3 that although \(\widehat{\mathcal S}_{\mathrm B}\) Code Structure uses fewer output tokens, synthesis round count is comparatively close. Program organization can therefore reduce the tokens required to repair the model each synthesis round. See Appendix~\ref{app:repair-controls} for details.

\subsection{RQ3: Effect of persistent NL on model learning}
\label{sec:nl-results}

\begin{figure}[h]
\centering
\includegraphics[width=0.82\linewidth]{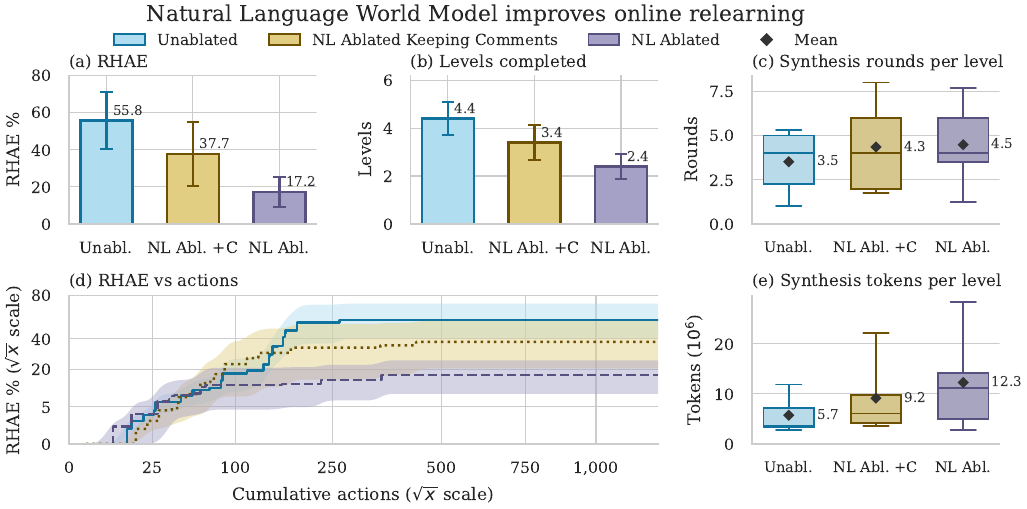}
\caption{\textbf{Model learning with and without persistent NL.} For graphs c and e, lower is better. For graphs a, b and d, higher is better. Detailed run values are provided in Appendix~\ref{app:nl-intervention}.}
\label{fig:rq3-hybrid}
\end{figure}

Compared to NL ablations, the complete system requires fewer actions, synthesis rounds, and tokens per level cleared on average during learning when object behaviors are poorly understood and reaches a higher RHAE (Figure~\ref{fig:rq3-hybrid}). Keeping code comments recovers part of this gap, but still requires more tokens and synthesis rounds per level than the complete system, most clearly in synthesis on the levels that all three conditions cleared. In the fixed-evidence RQ2 comparison, \(\widehat{\mathcal S}_{\mathrm B}\) Code Structure also has higher mean held-out accuracy than No NL. This suggests that persistent language helps synthesis interpret observations in addition to any improvement in action selection.

Both conditions encode histories and executable mechanics in Python. The full system's artifacts additionally retain the observations supporting a hypothesis, superseded explanations, and unresolved questions. These records preserve hypotheses when the evidence is insufficient. In the NL ablation traces, failures often concern sparsely tested behaviors with several plausible explanations. Retaining these qualitative explanations helps later synthesis revise an uncertain model without needing to reconstruct why its rules were introduced.

\section{Related Work}
\label{sec:related}

There is a vast amount of work on world models. We defer a broad summary of related areas to Appendix~\ref{app:related-work}. Successful recent systems on ARC-AGI-3 combine frontier models with graph-based exploration, executable world models, multi-agent workspaces, and continual agent programmatic harnesses \citep{rudakov2025graph,rodionov2026executableworldmodelsarcagi3,
sarafian2026workspace,lehmann2026tycho,skoutnev2026twin,karten2026continual,furgale2026nooa}, but lack formal approaches to efficient, information-informed exploration. OPINE-World uses a hybrid NL, Bayesian, and programmatic world model to guide exploration and revision, allowing a comparatively weak base model to greatly improve its capability in causal reasoning and world model building.

\section{Limitations}
\label{sec:limits}

We limit our evaluation to deterministic discrete ARC-AGI-3 environments, extension to stochastic or continuous domains would require significant extensions. Replay consistency also provides no guarantees for entirely unvisited objects. Appendix~\ref{app:limitations} discusses these limitations in detail.

\section{Conclusion}
\label{sec:conclusion}

We introduced OPINE-World, which combines Bayesian epistemic guidance, NL world models, and programmatic world models for online world-model learning. On ARC-AGI-3, we show that Bayesian guidance reduces catastrophic discovery failures, \(\widehat{\mathcal S}_{\mathrm B}\) Code Structure supports efficient repair, and an NL proto-representation reduces learning costs. On easier domains MiniGrid and Sokoban, OPINE-World also improves action efficiency over WorldCoder. These results support preserving a hybrid NL-Programmatic world model and using \(\widehat{\mathcal S}_{\mathrm B}\)for open-world problem solving.

\section*{Reproducibility Statement}
We provide source code for OPINE-World as supplementary material. Its README documents installation, how to fetch the public ARC-AGI-3 games, the Docker sandbox, and a launch script that runs games with the configuration used for our main results and ablations in RQ1 to RQ3. The models, reasoning effort, action budgets, and the constants of the learning loop and of \(\widehat{\mathcal S}_{\mathrm B}\) are listed in Appendix~\ref{app:experiments}, the verifier and the admission checks are specified in Appendix~\ref{app:rddl}, and scoring follows Appendix~\ref{app:scoring}. Because the agents are language models acting in a live environment, repeated runs vary in their action counts; the statistical tests in Appendix~\ref{app:experiments} account for this variation.

\bibliographystyle{iclr2027_conference}
\bibliography{iclr2027_conference}

@article{ahmed2025theorycoder,
  author = {Zergham Ahmed and Joshua~B. Tenenbaum and Christopher~J. Bates and Samuel~J. Gershman},
  title = {Synthesizing world models for bilevel planning},
  journal = {Transactions on Machine Learning Research},
  year = {2025},
  url = {https://openreview.net/forum?id=m9V4JHLJrD}
}

@inproceedings{alur2013sygus,
  author = {Rajeev Alur and Rastislav Bod{\'i}k and Garvit Juniwal and Milo M.~K. Martin and Mukund Raghothaman and Sanjit~A. Seshia and Rishabh Singh and Armando Solar-Lezama and Emina Torlak and Abhishek Udupa},
  title = {Syntax-guided synthesis},
  booktitle = {FMCAD},
  pages = {1--8},
  year = {2013}
}

@article{arcagi3,
  author = {{ARC Prize Foundation}},
  title = {{ARC-AGI-3}: A new challenge for frontier agentic intelligence},
  journal = {arXiv preprint arXiv:2603.24621},
  year = {2026},
  url = {https://arxiv.org/abs/2603.24621}
}

@misc{arcscorecards2026,
  author = {{ARC Prize Foundation}},
  title = {{ARC-AGI-3} scorecards},
  howpublished = {Documentation},
  year = {2026},
  url = {https://docs.arcprize.org/scorecards},
  note = {Accessed 21 September 2026}
}

@article{battaglia2018relational,
  author = {Peter~W. Battaglia and Jessica~B. Hamrick and Victor Bapst and Alvaro Sanchez-Gonzalez and Vinicius Zambaldi and Mateusz Malinowski and Andrea Tacchetti and David Raposo and Adam Santoro and others},
  title = {Relational inductive biases, deep learning, and graph networks},
  journal = {arXiv preprint arXiv:1806.01261},
  year = {2018}
}

@inproceedings{campbell2018cognitive,
  author = {G.~Ann Campbell},
  title = {Cognitive complexity: An overview and evaluation},
  booktitle = {TechDebt},
  pages = {57--58},
  year = {2018},
  doi = {10.1145/3194164.3194186}
}

@inproceedings{diuk2008oomdp,
  author = {Carlos Diuk and Andre Cohen and Michael~L. Littman},
  title = {An object-oriented representation for efficient reinforcement learning},
  booktitle = {ICML},
  pages = {240--247},
  year = {2008}
}

@article{furgale2026nooa,
  author = {Paul Furgale and Severin Klingler and James Nolan and Matt Staats and Di~Lorenzo, Gaia and Martinez~Abad, Elisa and Christian Sch{\"u}ller and Razvan Dinu and Alessio Devoto and Pascal Berard and Gal Kaplun and Elad Sarafian and Riccardo Roveri and Leon Derczynski and Ricardo Silveira Cabral},
  title = {{NVIDIA-labs OO Agents}: Native {Python} object-oriented agents},
  journal = {arXiv preprint arXiv:2607.20709},
  year = {2026},
  url = {https://arxiv.org/abs/2607.20709}
}

@book{gelman2013bayesian,
  author = {Andrew Gelman and John~B. Carlin and Hal~S. Stern and David~B. Dunson and Aki Vehtari and Donald~B. Rubin},
  title = {Bayesian Data Analysis},
  publisher = {Chapman and Hall/CRC},
  edition = {third},
  year = {2013}
}

@article{guestrin2003efficient,
  author = {Carlos Guestrin and Daphne Koller and Ronald Parr and Shobha Venkataraman},
  title = {Efficient solution algorithms for factored {MDP}s},
  journal = {Journal of Artificial Intelligence Research},
  volume = {19},
  pages = {399--468},
  year = {2003}
}

@inproceedings{ha2018worldmodels,
  author = {David Ha and J{\"u}rgen Schmidhuber},
  title = {Recurrent world models facilitate policy evolution},
  booktitle = {NeurIPS},
  pages = {2455--2467},
  year = {2018}
}

@article{hafner2025dreamerv3,
  author = {Danijar Hafner and Jurgis Pasukonis and Jimmy Ba and Timothy Lillicrap},
  title = {Mastering diverse control tasks through world models},
  journal = {Nature},
  volume = {640},
  number = {8059},
  pages = {647--653},
  year = {2025},
  doi = {10.1038/s41586-025-08744-2}
}

@article{hirschman1964paternity,
  author = {Albert~O. Hirschman},
  title = {The paternity of an index},
  journal = {American Economic Review},
  volume = {54},
  number = {5},
  pages = {761},
  year = {1964}
}

@inproceedings{houthooft2016vime,
  author = {Rein Houthooft and Xi Chen and Yan Duan and John Schulman and De~Turck, Filip and Pieter Abbeel},
  title = {{VIME}: Variational information maximizing exploration},
  booktitle = {NeurIPS},
  year = {2016}
}

@inproceedings{jha2010oracle,
  author = {Susmit Jha and Sumit Gulwani and Sanjit~A. Seshia and Ashish Tiwari},
  title = {Oracle-guided component-based program synthesis},
  booktitle = {ICSE},
  pages = {215--224},
  year = {2010}
}

@article{karten2026continual,
  author = {Seth Karten and Joel Zhang and Upaa, Jr., Tersoo and Ruirong Feng and Wenzhe Li and Chengshuai Shi and Chi Jin and Kiran Vodrahalli},
  title = {Continual harness: Online adaptation for self-improving foundation agents},
  journal = {arXiv preprint arXiv:2605.09998},
  year = {2026},
  url = {https://arxiv.org/abs/2605.09998}
}

@inproceedings{kipf2020contrastive,
  author = {Thomas Kipf and Elise van~der Pol and Max Welling},
  title = {Contrastive learning of structured world models},
  booktitle = {ICLR},
  year = {2020},
  url = {https://openreview.net/forum?id=H1gax6VtDB}
}

@misc{lacchia2023radon,
  author = {Michele Lacchia},
  title = {{Radon 6.0.1} documentation},
  howpublished = {Software documentation},
  year = {2023},
  url = {https://radon.readthedocs.io/en/stable/},
  note = {Accessed 21 September 2026}
}

@article{lehmann2026tycho,
  author = {Jens Lehmann and Andrei Aioanei and Sahar Vahdati},
  title = {{Tycho}: Active abstraction with programmatic world models for {ARC-AGI-3}},
  journal = {arXiv preprint arXiv:2607.28287},
  year = {2026},
  url = {https://arxiv.org/abs/2607.28287}
}

@article{levy2025worldllm,
  author = {Guillaume Levy and C{\'e}dric Colas and Pierre-Yves Oudeyer and Thomas Carta and Cl{\'e}ment Romac},
  title = {{WorldLLM}: Improving {LLM}s' world modeling using curiosity-driven theory-making},
  journal = {arXiv preprint arXiv:2506.06725},
  year = {2025},
  url = {https://arxiv.org/abs/2506.06725}
}

@article{lindley1956measure,
  author = {Dennis~V. Lindley},
  title = {On a measure of the information provided by an experiment},
  journal = {Annals of Mathematical Statistics},
  volume = {27},
  number = {4},
  pages = {986--1005},
  year = {1956}
}

@article{mccabe1976complexity,
  author = {Thomas~J. McCabe},
  title = {A complexity measure},
  journal = {IEEE Transactions on Software Engineering},
  volume = {SE-2},
  number = {4},
  pages = {308--320},
  year = {1976},
  doi = {10.1109/TSE.1976.233837}
}

@phdthesis{mccallum1996reinforcement,
  author = {Andrew~K. McCallum},
  title = {Reinforcement learning with selective perception and hidden state},
  school = {University of Rochester},
  year = {1996}
}

@inproceedings{pathak2017curiosity,
  author = {Deepak Pathak and Pulkit Agrawal and Alexei~A. Efros and Trevor Darrell},
  title = {Curiosity-driven exploration by self-supervised prediction},
  booktitle = {ICML},
  pages = {2778--2787},
  year = {2017}
}

@inproceedings{poeworld2025,
  author = {Wasu~Top Piriyakulkij and Yichao Liang and Hao Tang and Adrian Weller and Marta Kryven and Kevin Ellis},
  title = {{PoE-World}: Compositional world modeling with products of programmatic experts},
  booktitle = {NeurIPS},
  year = {2025},
  url = {https://arxiv.org/abs/2505.10819}
}

@inproceedings{rodionov2026executableworldmodelsarcagi3,
  author = {Sergey Rodionov},
  title = {Executable world models for {ARC-AGI-3} in the era of coding agents},
  booktitle = {AGI},
  series = {Lecture Notes in Computer Science},
  volume = {16855},
  pages = {198--210},
  publisher = {Springer},
  year = {2026},
  doi = {10.1007/978-3-032-33195-3_15},
  url = {https://arxiv.org/abs/2605.05138}
}

@unpublished{sanner2010rddl,
  author = {Scott Sanner},
  title = {Relational dynamic influence diagram language ({RDDL}): Language description},
  note = {Unpublished manuscript, Australian National University},
  year = {2010},
  url = {http://users.cecs.anu.edu.au/~ssanner/IPPC_2011/RDDL.pdf}
}

@article{sarafian2026workspace,
  author = {Elad Sarafian and Gal Kaplun and Ron Banner and Daniel Soudry and Boris Ginsburg},
  title = {Workspace optimization: How to train your agent},
  journal = {arXiv preprint arXiv:2605.09650},
  year = {2026},
  url = {https://arxiv.org/abs/2605.09650}
}

@inproceedings{sekar2020planning,
  author = {Ramanan Sekar and Oleh Rybkin and Kostas Daniilidis and Pieter Abbeel and Danijar Hafner and Deepak Pathak},
  title = {Planning to explore via self-supervised world models},
  booktitle = {ICML},
  pages = {8583--8592},
  year = {2020}
}

@article{shannon1948,
  author = {Claude~E. Shannon},
  title = {A mathematical theory of communication},
  journal = {Bell System Technical Journal},
  volume = {27},
  pages = {379--423, 623--656},
  year = {1948}
}

@article{skoutnev2026twin,
  author = {Alexy Skoutnev and Kirill Acharya and Gaston Longhitano and Madeleine Udell and Kevin Ellis and Iddo Drori},
  title = {{Twin}: Playing an unknown game with a test-time digital twin},
  journal = {arXiv preprint arXiv:2608.14490},
  year = {2026},
  url = {https://arxiv.org/abs/2608.14490}
}

@inproceedings{solarlezama2006sketching,
  author = {Armando Solar-Lezama and Liviu Tancau and Rastislav Bod{\'i}k and Sanjit~A. Seshia and Vijay~A. Saraswat},
  title = {Combinatorial sketching for finite programs},
  booktitle = {ASPLOS},
  pages = {404--415},
  year = {2006}
}

@inproceedings{strehl2007efficient,
  author = {Alexander~L. Strehl and Carlos Diuk and Michael~L. Littman},
  title = {Efficient structure learning in factored-state {MDP}s},
  booktitle = {AAAI},
  pages = {645--650},
  year = {2007}
}

@article{sutton1991dyna,
  author = {Richard~S. Sutton},
  title = {Dyna, an integrated architecture for learning, planning, and reacting},
  journal = {ACM SIGART Bulletin},
  volume = {2},
  number = {4},
  pages = {160--163},
  year = {1991}
}

@inproceedings{tang2024worldcoder,
  author = {Hao Tang and Darren Key and Kevin Ellis},
  title = {{WorldCoder}, a model-based {LLM} agent: Building world models by writing code and interacting with the environment},
  booktitle = {NeurIPS},
  year = {2024}
}

@article{rudakov2025graph,
  author = {Evgenii Rudakov and Jonathan Shock and Benjamin~Ultan Cowley},
  title = {Graph-based exploration for {ARC-AGI-3} interactive reasoning tasks},
  journal = {arXiv preprint arXiv:2512.24156},
  year = {2025},
  url = {https://arxiv.org/abs/2512.24156},
  note = {AAAI 2026 Workshop on AI for Scientific Research}
}

@inproceedings{
liu2026natural,
title={Natural Language {PDDL} ({NL}-{PDDL}) for Open-world Goal-oriented Commonsense Regression Planning in Embodied {AI}},
author={Xiaotian Liu and Armin Toroghi and Jiazhou Liang and David Courtis and Ruiwen Li and Ali Pesaranghader and Jaehong Kim and Tanmana Sadhu and Hyejeong Jeon and Scott Sanner},
booktitle={The Fourteenth International Conference on Learning Representations},
year={2026},
url={https://openreview.net/forum?id=kWCNhRdcDI}
}

\appendix

\section{Worked Example: \texorpdfstring{\code{ft09}}{ft09}}
\label{app:ft09-example}

\begin{runningexample}{\code{ft09} Example Domain}
\label{ex:ft09-problem}
Consider this elementary ARC-AGI-3 problem.  A set of \textcolor{opineblue}{answer cells} neighbors \textcolor{opineorange}{goal squares}.  Clicking an \textcolor{opineblue}{answer cell} advances its color through the ordered \textcolor{opinegreen}{palette}; in later levels, the click may also change nearby cells.  Each \textcolor{opinemagenta}{directional position} on a neighboring \textcolor{opineorange}{goal square} specifies the target color of the \textcolor{opineblue}{answer cell} in that relative position.  The goal is to make every \textcolor{opineblue}{answer cell} match its directional \textcolor{opineorange}{goal square} colour.

Let \(\textcolor{opineblue}{\mathcal C}\) be the answer cells, \(\textcolor{opineorange}{\mathcal G}\) the goal squares, \(\textcolor{opinegreen}{\mathcal P}\) the ordered palette, \(\textcolor{opinemagenta}{\gamma(c)}\) the target decoded from \(\mathcal G\), and \(B(c,d)\) indicate that clicking \(c\) also changes \(d\).  Writing \(\sym{CLICK}(c)\) for a coordinate click that hits \(c\):
\begin{align*}
\textbf{State:}\quad
  &\mathcal S=\{s:\mathcal C\!\to\!\mathcal P\},
    \qquad s(c)=\sym{color}(c),\\
\textbf{Action:}\quad
  &\mathcal A=\{\sym{CLICK}(x,y):x,y\in\{0,\ldots,63\}\},
    \qquad |\mathcal A|=64^2=4096,\\
\textbf{Transition:}\quad
  &T(s,\sym{CLICK}(c))(d)=
    \begin{cases}
      \sym{next}_{\mathcal P}(s(d)),&d=c\ \lor\ B(c,d),\\
      s(d),&\text{otherwise},
    \end{cases}\\
\textbf{Reward:}\quad
  &R(s,a,s')=\mathbf 1\!\left[\forall c\in\mathcal C,
    \ s'(c)=\gamma(c)\right].
\end{align*}
\end{runningexample}

Figure~\ref{fig:ft09-pipeline} in Section~\ref{sec:model-learning-loop} follows one pass of the learning loop on Level 6 of this game, where a click also flips the tile above.

\paragraph{Factored dynamics.} In this example, color is a state fluent, the grounded click is an action fluent, and the palette, key pixels, and cell geometry are fixed within a level. The transition can be written as a deterministic CPF (Appendix~\ref{app:cpfs}):
\begin{align}
  \sym{advance}(c;s,a)
  &\equiv \sym{click}(c)\lor\exists d\,[\sym{click}(d)\wedge\sym{brushes}(d,c;s)],
  \label{eq:ft09-advance}\\
  \sym{color}'(c)
  &=\begin{cases}
    \sym{next}_{\sym{palette}}(\sym{color}(c)),&\sym{advance}(c;s,a),\\
    \sym{color}(c),&\text{otherwise}.
  \end{cases}
  \label{eq:ft09-color-update}
\end{align}
Its parents include the current color, ordered palette, click target, and brush relation. The derived fluent \(\sym{cell_brush_arity}(d)=|\{c:\sym{brushes}(d,c;s)\}|\) counts how many other cells change; the binary relation identifies which cells change. The same neighborhood geometry determines \(\sym{required_color}(c)\), used by the reward model. These dependencies localize prediction errors while allowing Python to compute coupled effects jointly.

\subsection{Qualitative and programmatic reward representations}
\label{app:ft09-reward}
\begin{runningexamplecontinued}{Reward}
\label{ex:ft09-reward-parallel}
\noindent
\begin{minipage}[t]{0.43\linewidth}
\vspace{0pt}
\centering
\textbf{\(\widehat R_{\mathrm{NL}}\): qualitative}\par
\begin{lstlisting}[style=conceptmap,basicstyle=\ttfamily\fontsize{5.9}{6.5}\selectfont]
<O>Each key assigns a required color to an answer</O>
<O>cell.</O>

<B>A cell is satisfied when its color equals its</B>
<B>required color </B>and<A> there is no submit action.</A>

<G>Reward is 1 exactly when every answer cell</G>
<G>is satisfied at once.</G>

<M>Hypothesis: On three-color levels, gray may</M>
<M>be a neutral target.</M>

<L>Heuristic: Try high arity keys first.</L>
\end{lstlisting}
\end{minipage}\hfill
\begin{minipage}[t]{0.55\linewidth}
\vspace{0pt}
\centering
\textbf{\(\widehat R_{\mathrm P}\): quantitative}\par
\begin{lstlisting}[style=conceptmap,basicstyle=\ttfamily\fontsize{5.8}{6.4}\selectfont]
@fluent
<O>def required_color(c, state):</O>
<O>    return unique_value(key_demands(c, state))</O>

@fluent
<B>def satisfied(c, state):</B>
<B>    return color(c, state) == required_color(c, state)</B>

def reward_function(state, action, _):
<A>    after_click = apply_click(state, action)</A>
<G>    solved = all(satisfied(c, after_click)</G>
<G>                 for c in answer_cells(after_click))</G>
<G>    return float(solved)</G>
\end{lstlisting}
\end{minipage}
In \code{ft09}, \(r^{0}=0\), \(r^{+}=1\), and the inferred completion predicate is
\begin{equation}
\widehat G(\widehat s')=
\mathbf 1\!\left[
  \forall c\in\widehat{\mathcal C}(\widehat s'),\;
  \sym{color}(c;\widehat s')
  =\sym{required_color}(c;\widehat s')
\right].
\label{eq:ft09-goal-predicate}
\end{equation}
Notice that NL and Python contain different abstraction levels. Keeping both preserves compact, nuance-rich explanations without weakening executability.
\end{runningexamplecontinued}

\section{RDDL Formalization and Diagnostic Details}
\label{app:rddl}

For compactness, this appendix writes \(\widehat T\), \(\widehat R\), and \(\widehat P\) for the programmatic components \(\widehat T_{\mathrm P}\), \(\widehat R_{\mathrm P}\), and \(\widehat P\) introduced in the main text. We use the vocabulary of RDDL and factored MDPs to describe their object-based state and transition structure \citep{sanner2010rddl,guestrin2003efficient}.

\subsection{Observation abstraction and grounded fluent states}
\label{app:rddl-state}

The environment emits an observation $\omega_t$. Let $f$ be the rendering map and let $f^{-1}$ denote the learned or supplied abstraction:
\begin{equation}
  \omega_t=f(s_t),
  \qquad
  \widehat s_t=f^{-1}(\widehat s_{t-1}, \omega_t).
  \label{eq:observation-map}
\end{equation}
Equation~\ref{eq:observation-map} uses $f^{-1}$ to denote a parser. The extracted state is amodal, subject to the LLM priors, and can infer occluded objects. That is, a generated engine may maintain history $\widehat s_{t-1}$ which includes the extracted record and the retained history. Replay requires the current values of $\widehat s_t$ to contain the information needed to predict the successor. This construction covers hidden state that can be recovered deterministically from the observed trajectory. A model of irreducible ambiguity, such as random effects, would require a distribution over possible states, which is out of the scope of this system.

Throughout, $\widehat s_t$ includes any relevant history reconstructed from replay, so the Markov transition $\widehat T(\widehat s_t,a_t)$ conditions on this augmented state.

Each extracted object record contains an identity, inferred type, and fields such as position, pixels, color, rotation, and visibility. Matching records across observations distinguishes changes to an existing object from appearances and disappearances.

\begin{table}[H]
\caption{\textbf{RDDL notation.} Symbols used in the method.}
\label{tab:rddl-notation}
\centering
\small
\begin{tabular}{@{}L{0.18\linewidth}L{0.72\linewidth}@{}}
\toprule
Symbol & Meaning \\
\midrule
$\Types,\Obj$ & object types and their concrete instances \\
$\SF$ & state properties written with object variables \\
$\AF$ & action indicators and the action $a_t$ executed at step $t$ \\
$\NF$ & level information that actions don't alter, such as walls \\
$\IF_t$ & values produced at step $t$ by derived functions of the current state and action \\
$X(\bar o)$ & schema $X(\bar z)$ grounded with a legal object tuple $\bar o$ \\
$\Val_X$ & allowed values of state property $X$ \\
$\Pa_X$ & selects the inputs used by the update rule for $X$ \\
$u_t^X(\bar o)$ & values of those selected inputs for objects $\bar o$ at step $t$ \\
$\tau(o)$ & assigns object $o$ to a type whose members share an update-rule template \\
$\Data_t$ & observed transition and reward history \\
\bottomrule
\end{tabular}
\end{table}

A lifted schema $X(\bar z)$ has $k$ typed object arguments $\bar z=(z_1,\ldots,z_k)$. Grounding it with compatible objects $\bar o=(o_1,\ldots,o_k)$ produces $X(\bar o)$. The state assigns one value to each legal grounding:
\begin{equation}
  s_t=
  \{X_t(\bar o)=v_t^{X,\bar o}\mid
    X\in\SF,\ \bar o\in\mathrm{Ground}(X),\
    v_t^{X,\bar o}\in\Val_X\}.
  \label{eq:full-ground-state}
\end{equation}
In Equation~\ref{eq:full-ground-state}, $\mathrm{Ground}(X)$ is the set of legal object tuples for schema $X$, $\Val_X$ is its value domain, and $v_t^{X,\bar o}$ is the value assigned to grounding $X(\bar o)$ at step $t$. For example, $x(p_1)$, $\sym{shape}(k_1)$, and $\sym{locked}(d_1)$ are grounded fluents. The implementation stores fluents with a shared object argument in the same object record. This record is a storage representation of the grounded assignment.

\begin{runningexamplecontinued}{Grounded states}
\label{ex:ls20-grounded}
Suppose the player is at cell $(2,3)$ and action 4 means move right. An open destination gives
\[
  x_t(p_1)=2,\ y_t(p_1)=3
  \quad\longrightarrow\quad
  x_{t+1}(p_1)=3,\ y_{t+1}(p_1)=3.
\]
If a wall occupies the destination, the same action leaves $x_{t+1}(p_1)=2$. Both transitions assign values to the schemas $x(p)$ and $y(p)$.
\end{runningexamplecontinued}

\begin{table}[H]
\caption{\textbf{Fluent vocabulary for the appendix micro-example.} The examples give simple names to roles inferred in an ARC-AGI-3 game environment named \code{ls20}.}
\label{tab:rddl-vocabulary-app}
\centering
\small
\begin{tabular}{@{}L{0.18\linewidth}L{0.72\linewidth}@{}}
\toprule
Category & Meaning in the world-model setting \\
\midrule
$\Types$ & object types or shared templates, such as player, key, rotator, door, and wall \\
$\Obj$ & objects in the current level, such as player $p_1$, key $k_1$, rotator $r_1$, and door $d_1$ \\
$\NF$ & level information that actions do not alter, such as wall cells, fixed object locations, and a door's target key shape \\
$\SF$ & state properties written with object variables, such as $x(p)$, $y(p)$, $\sym{shape}(k)$, and $\sym{locked}(d)$ \\
$\AF$ & action indicators such as $\sym{move_right}(p)$ \\
$\IF$ & derived functions such as $\sym{blocked_right}(p)$ and $\sym{at_rotator}(p,r)$ \\
CPF & update rule for one next-state fluent, such as the rule for $x'(p)$ \\
\bottomrule
\end{tabular}
\end{table}

\subsection{Factored dynamics and conditional probability functions}
\label{app:cpfs}

A conditional probability function (CPF) specifies each next-state fluent from a parent set. Non-fluents describe level information that an action does not alter. In the running example, this includes the wall map and the locations of the fixed rotator and door. A CPF can read this information when predicting what changes.

RDDL intermediate fluents are implemented here as derived functions of the current state and action. Their values are recalculated before the model predicts the next state rather than stored across steps. For example, $\sym{blocked_right}_t(p)$ returns whether a wall occupies the cell to the player's right. The position rule uses this result to decide whether movement succeeds. Model-declared \code{@fluent} functions expose derived object properties and relations to the Bayesian diagnostics, subject to the admission checks in Appendix~\ref{app:refinement}.

For a next-state fluent $X'(\bar z)$ and a grounding $\bar o$, the CPF has the form
\begin{equation}
  P\!\left(X_{t+1}(\bar o)\mid\Pa_X(\bar o;s_t,a_t,\NF,\IF_t)\right).
  \label{eq:general-cpf}
\end{equation}
Equation~\ref{eq:general-cpf} uses $\Pa_X$ for the parent or context set of $X'(\bar o)$. It selects the subset of current state fluents, action fluents, non-fluents, and intermediate fluents relevant to predicting this next fluent. The experiments are deterministic, which reduces this conditional distribution to a point mass:
\begin{equation}
  P\!\left(X_{t+1}(\bar o)=v\mid s_t,a_t\right)
  =\mathbf 1\!\left\{v=f_X\!\left(\bar o;u_t^X(\bar o)\right)\right\}.
  \label{eq:deterministic-cpf}
\end{equation}
In Equation~\ref{eq:deterministic-cpf}, $v$ is a candidate next value and $\mathbf 1\{\cdot\}$ equals one when its condition is true and zero otherwise. The equation assigns all probability to the value returned by $f_X$.

\subsubsection{A tiny lifted model}
\label{app:lifted-ls20-model}

The following lifted CPFs serve as an example domain.
\begin{lstlisting}
blocked_right(p) :=
  wall_at(x(p)+1, y(p))

x'(p) =
  if move_right(p) and not blocked_right(p)
  then x(p)+1
  else x(p)

y'(p) =
  y(p)

at_rotator(p,r) :=
  x(p)=rotator_x(r) and y(p)=rotator_y(r)

shape'(k) =
  if exists p,r: at_rotator(p,r)
  then rotate(shape(k))
  else shape(k)

key_fits(k,d) :=
  shape(k)=target_shape(d)

at_door(p,d) :=
  x(p)=door_x(d) and y(p)=door_y(d)

locked'(d) =
  if exists k: key_fits(k,d)
  then false
  else locked(d)
\end{lstlisting}

First, the CPFs are \emph{local}. The update for $x'(p)$ does not need the complete frame. It needs the current position, the relevant action, and a local obstacle condition summarized by $\sym{blocked_right}(p)$. Second, the CPFs have \emph{semantic bodies}: the learner must infer which fluent an input changes, by how much, and under which conditions. Third, the CPFs are \emph{lifted}: the rule for $x'(p)$ is written once over a player variable $p$, then grounded for $p_1$. The rules for $\sym{shape}'(k)$ and $\sym{locked}'(d)$ are likewise written over key and door variables. A learned template can therefore be reused for compatible objects in later levels.

\subsubsection{Locality and CPF parent error}
\label{app:cpf-parent-error}

Technically, a CPF may condition on the complete state:
\begin{equation}
  X_{t+1}(\bar o)=f_X(\bar o;s_t,a_t).
  \label{eq:global-cpf}
\end{equation}
Equation~\ref{eq:global-cpf} is formally valid, but it does not identify which parts of the state support reuse or diagnosis. A useful model identifies a compact parent assignment for each next fluent:
\begin{equation}
  u_t^X(\bar o)\subseteq s_t\cup a_t\cup\NF\cup\IF_t.
  \label{eq:local-context}
\end{equation}
Equation~\ref{eq:local-context} restricts this assignment to values available from the current state, action, non-fluents, and derived functions. For the player's horizontal coordinate, a sufficient context contains the current position, the action, and the binary wall test in the \hyperref[ex:ls20-grounded]{grounded transition example}. A mixed empirical group can therefore indicate that the proposed context is too coarse. If this test is omitted, players governed by the same lifted template can show both movement and no-change effects. Adding the test splits those effects into predictable contexts.

This is the RDDL interpretation of a missing-parent error. The type and action may be correct, while the context supplied to the CPF omits a fluent or derived relation that changes the outcome. The observable ledger and declared-fluent registry test this explanation by measuring whether a candidate parent makes the resulting groups more predictable.

\begin{runningexamplecontinued}{Mixed groups mean missing CPF parents}
Suppose the group only indexes
\[
  j=(\sym{player},\sym{right},u_0)
\]
and records
\[
  C(j,\sym{x_change})=6,
  \qquad
  C(j,\sym{no_change})=4.
\]
This need not mean that movement is stochastic (and we assume no stochasticity). The context $u_0$ may be missing $\sym{blocked_right}(p)$. Adding that parent gives
\[
\begin{array}{lll}
u=\sym{open_right}: & C(\sym{x_change})=6, & \text{move succeeds},\\
u=\sym{wall_right}: & C(\sym{no_change})=4, & \text{move is blocked}.
\end{array}
\]
The two contexts now predict separate outcomes, as expected from a deterministic movement CPF.
\end{runningexamplecontinued}

\subsubsection{CPF body and action-semantic error}
\label{app:cpf-semantic-error}

Even with the correct local parents, the CPF body may still be wrong. This part of world-model learning identifies the action semantics and the update equation. For the simplified \code{ls20} example, a learner may correctly identify position, action 4, and the wall test as parents, but it must still infer that an open destination adds one to $x$ and a blocked destination leaves $x$ unchanged.

The executed input $a_t$ is observed, but its effect is not known in advance. In RDDL terms, the learner knows which action assignment occurred, but it must infer which next fluents depend on that action and how. If the parent set is adequate but an exact next value is wrong, the error lies in the CPF body rather than the parent selection. CEGIS tests this distinction because it compares the values predicted by the executable body, while the effect diagnostic records a coarser summary.

\begin{runningexamplecontinued}{Learning the CPF body}
Suppose the learner has correctly identified
\[
  \Pa_x(p)=\{x_t(p),y_t(p),\sym{move_right}_t(p),\sym{blocked_right}_t(p)\}.
\]
It must still learn whether action 4 changes $x$ by $+1$, by $-1$, or not at all. The correct deterministic body is
\[
  f_x(p;\Pa_x)=
  \begin{cases}
    x_t(p)+1,&\sym{move_right}_t(p)\wedge\neg\sym{blocked_right}_t(p),\\
    x_t(p),&\text{otherwise}.
  \end{cases}
\]
\end{runningexamplecontinued}

\subsubsection{Liftedness and typing error}
\label{app:typing-error}

A CPF is written with variables and then grounded to objects. The lifted template
\begin{equation}
  x'(p)=f_x(p;s,a)
  \label{eq:lifted-template}
\end{equation}
applies to each compatible player $p$. More generally, a property update may be shared by a type or concept $\tau$:
\begin{equation}
  b'(o)=f_{\tau,b}(o;s,a),
  \qquad \tau(o)=\tau.
  \label{eq:typed-template}
\end{equation}
Typing defines which objects share a lifted CPF template. If types are known, they are declared RDDL object categories, called sorts. If types are uncertain, $\tau(o)$ may be treated as an attribute or learned concept that selects the update rule. 

Liftedness creates a separate error mode. If a template is shared across objects that should have different dynamics, one empirical group can mix incompatible effects even when the local context is adequate. For example, a moving block and a static wall should not share a position-update template. The repair is to split or revise the type used to select the CPF template. Parent error and typing error can initially produce the same mixed group. The system therefore tests candidate contexts and verifies the resulting program by replay before accepting a structural revision.

\begin{runningexamplecontinued}{Lifted reuse}
If the agent learns that a player moves right when the destination is open, it is represented as a lifted rule:
\[
  x'(p)=x(p)+1
  \quad\text{if}\quad
  \sym{move_right}(p)\wedge\neg\sym{blocked_right}(p).
\]
The same rule can be grounded on each later object assigned the player type.
\end{runningexamplecontinued}

\subsubsection{Grounding gives a factored dynamic model}

Once a lifted RDDL model is grounded over a finite object set, each grounded fluent becomes a node in a two-slice dynamic Bayesian network, a graph connecting current-state variables to next-state variables. Grounding the CPFs gives the factored transition
\begin{equation}
  P(s_{t+1}\mid s_t,a_t)
  =\prod_{X\in\SF}
   \prod_{\bar o\in\mathrm{Ground}(X)}
   P\!\left(X_{t+1}(\bar o)\mid
     \Pa_X(\bar o;s_t,a_t)\right).
  \label{eq:factorization}
\end{equation}
Each factor is a point mass in the deterministic experiments.

Current fluents, the action, and fixed instance facts determine intermediate fluents and a local parent assignment. The type selects a shared CPF template, which produces one next-state fluent. Grounding this pattern for each legal object and fluent yields Equation~\ref{eq:factorization}. The synthesized Python evaluates these factors together inside \code{transition_function}.

\subsection{Effect signatures}
\label{app:rows}

The exact CPF context depends on a fluent family $b$. A detailed diagnostic could therefore use
\begin{equation}
  j_t^{i,b}=(\tau_t^i,b,a_t,c_t^{i,b}).
  \label{eq:cpf-row}
\end{equation}
Equation~\ref{eq:cpf-row} gives the fluent-specific group suggested by the CPF formalism. A group in this case is simply a unique permutation of its contents. The implementation uses the coarser object group $j_t^i=(\tau_t^i,a_t,c_t^i)$ of Section~\ref{sec:eta} and collects changed fluent families into one effect signature. This approximation reduces the number of groups and applies the same diagnostic to births, disappearances, movement, visibility, rotation, and pixel changes. The baseline $c_t^i$ is the sorted list of visible-neighbor types within Chebyshev radius $r_c$. Values from committed $\xi$ features are appended to this list. Python synthesis receives the complete recorded states and may introduce dependencies that are absent from $j_t^i$. The verifier then evaluates their full predicted values.

For a paired object $i$, define
\begin{equation}
  \Delta_t^i=\{b:b_t(o_i)\neq b_{t+1}(o_i)\},
  \qquad e_t^i=\rho(\Delta_t^i)\in\Effects.
  \label{eq:app-signature}
\end{equation}
Equation~\ref{eq:app-signature} maps the set of changed fields $\Delta_t^i$ to the categorical effect $e_t^i$. The implementation excludes size, display-coordinate, name, and tag metadata before constructing $\Delta_t^i$. It sorts the remaining field names and adds $\sym{no_change}$, $\sym{born}$, and $\sym{gone}$ categories. Object matching first uses name and position and then pairs remaining records with the same name.

\subsection{Bayesian uncertainty and exploration}
\label{app:eta}

The belief-state example in Section~\ref{sec:natural-language-world-model} is a schematic abstraction. Its displayed property and relation probabilities summarize uncertainty whose underlying attribution is lifted over objects, actions, contexts, and observed effects. The implementation records evidence for these effect classes and grounds model-declared fluents on applicable objects; the illustration omits this indexing and uses illustrative values rather than literal estimator outputs.

Let $C_t(j,e)$ count signatures in group $j$ and let $n_j=\sum_eC_t(j,e)$. A repeated $(j,e)$ pair contributes at most once per transition, which limits multiplicity from repeated instances. The global alphabet size is
\begin{equation}
  m=\max\left(m_{\min},
    \left|\bigcup_j\{e:C_t(j,e)>0\}\right|\right).
  \label{eq:alphabet}
\end{equation}
Equation~\ref{eq:alphabet} counts effect labels observed in any group and applies the lower bound $m_{\min}$, which keeps the normalized entropy defined before the ledger has observed multiple labels. For $q_j(e)=P(e_t^i=e\mid j)$, we place a symmetric Dirichlet prior on the categorical effect probabilities and update it with the recorded counts \citep{gelman2013bayesian}:
\begin{equation}
\begin{split}
  q_j&\sim\Dir(\alpha_0\mathbf 1_m),\\
  q_j\mid\Data_t&\sim
    \Dir(\alpha_0\mathbf 1_m+\mathbf c_j),\\
  \widehat q_j(e)&=
    \frac{\alpha_0+C_t(j,e)}{m\alpha_0+n_j}.
  \label{eq:dirichlet-row}
\end{split}
\end{equation}
In Equation~\ref{eq:dirichlet-row}, $q_j$ is the group's vector of effect probabilities, $\mathbf c_j$ is its count vector, $\mathbf 1_m$ is a vector of $m$ ones, and $\Dir$ denotes the Dirichlet distribution. The first line states the symmetric prior, the second updates it with recorded counts, and the third gives the probability used by the diagnostic. Unobserved effect categories in a group retain predictive mass $\alpha_0/(m\alpha_0+n_j)$. The normalized group and aggregate uncertainties use Shannon entropy \citep{shannon1948}:
\begin{equation}
  \eta_j=\frac{\Entropy(\widehat q_j)}{\log m},
  \qquad
  \eta_t=\frac{\sum_jn_j\eta_j}{\sum_jn_j}.
  \label{eq:app-eta}
\end{equation}
Equation~\ref{eq:app-eta} defines uncertainty for a group and its count-weighted aggregate over the ledger. Archived diagnostics label $\eta_t$ as ontology error. We use aggregate uncertainty because the statistic also responds to missing parents, extraction errors, and hidden state. The diagnostic report orders groups by $n_j\eta_j$. A group can have high entropy because its evidence is sparse or because its recorded outcomes remain mixed. Multiplication by $n_j$ reduces the priority of small groups and retains well-supported mixed groups. Neither entropy nor this ordering determines the cause. A proposed refinement is required to test a missing parent or an object type that combines distinct dynamics. Object-pairing error and hidden state can produce the same observations without being resolved by such a refinement.

The implementation also evaluates a finite set $\mathcal C_t$ of type merges and composite-object candidates. If $\mathcal R$ denotes the current type assignment, it reports
\begin{equation}
  \eta_t^\star=\min_{\mathcal R'\in\{\mathcal R\}\cup\mathcal C_t}
    \eta_t(\mathcal R'),
  \qquad
  \Delta\eta_t^\star=\eta_t(\mathcal R)-\eta_t^\star.
  \label{eq:eta-star}
\end{equation}
Equation~\ref{eq:eta-star} selects the lowest aggregate uncertainty among the current representation and the enumerated candidates. The reduction $\Delta\eta_t^\star$ can underestimate the reduction available from a representation that the candidate generator did not enumerate.


Actor-facing guidance has two parts. The first we consider an ``exploration matrix'', which has one entry for each object type and action $(\tau,a)$. For an entry with $n_{\tau,a}$ observations, of which $s_{\tau,a}$ show the most common effect, we place a uniform prior on the probability of the most common effect and update it with these counts. This gives a $\mathrm{Beta}(1+s_{\tau,a},\,1+n_{\tau,a}-s_{\tau,a})$ posterior. Entries are ordered by one minus a sample from this posterior. The matrix shows the actor which object--action effects are under-sampled or still mixed, including those whose confounder is unknown. This points towards objects of high complexity.

The second part tells the actor which mixed effects the model accounts for. An admitted model-declared fluent $\zeta(o,s)$ refines a mixed action--context class into bins indexed by its value $v$, with evidence shared across objects of the same type within each action--context--value bin. For a refined bin with $n_{j,v}$ observations and modal-outcome count $s_{j,v}$, its uncertainty is
\begin{equation}
  \text{Refined-bin uncertainty}=1-\frac{s_{j,v}}{n_{j,v}+1}.
  \label{eq:app-refined-uncertainty}
\end{equation}
This value approaches zero after a few consistent observations, as expected in a deterministic environment. When the bins of a mixed entry have low uncertainty, the actor can treat that entry as understood and spend its actions elsewhere. Fluent values observed without interaction evidence are listed as untested, which gives the actor unknown states to probe. Together, the two parts direct exploration toward effects that are complex or poorly understood.

\begin{runningexamplecontinued}{From effects to exploration priority}
The open and blocked transitions produce
\[
\begin{array}{ll}
j_{\rm open}=(\sym{player},4,c_{\rm open}), & e_{\rm open}=\sym{x_change},\\
j_{\rm wall}=(\sym{player},4,c_{\rm wall}), & e_{\rm wall}=\sym{no_change}.
\end{array}
\]
Once the fluent $\sym{blocked_right}$ is admitted, open and blocked contexts form separate bins. With $n$ observations of the same effect in either bin, Equation~\ref{eq:app-refined-uncertainty} gives uncertainty $1/(n+1)\to0$. Once both bins have low uncertainty, the actor treats the mixed $(\sym{player},4)$ entry as understood and explores elsewhere.
\end{runningexamplecontinued}

\subsection{Context refinements and declared fluents}
\label{app:refinement}

The $\xi$ ledger scores the $K_\xi$ groups with largest $n_j\eta_j$. Candidate features include varying target fields, a bounded target-pixel hash, neighbor types at observed offsets, neighbor multisets at radii in $\mathcal R_\xi$, and within-object click offsets for the pointer action. Generic target-field candidates exclude position, raw pixels, collision flags, and display metadata because those values can create instance-specific partitions.

For candidate $\xi$, write $j_v$ for the subgroup with $\xi=v$. The gate is
\begin{equation}
\begin{gathered}
  \eta_{j\mid\xi}=\sum_v\frac{n_{j_v}}{n_j}\eta_{j_v},
  \qquad \Delta\eta(j,\xi)=\eta_j-\eta_{j\mid\xi},\\
  \Delta\eta(j,\xi)\geq\delta_\xi,
  \qquad
  \exists v:\ n_{j_v}\geq n_\xi\ \land\
  \max_e C_t(j_v,e)/n_{j_v}\geq p_\xi.
  \label{eq:xi-gate}
\end{gathered}
\end{equation}
The threshold $\delta_\xi$ requires a minimum uncertainty reduction. The support threshold $n_\xi$ requires enough samples in at least one subgroup, and $p_\xi$ requires that subgroup to have a sufficiently common effect. The ledger supplies the candidate features and their statistics to the synthesizer. If the synthesizer proposes a feature, the engine re-evaluates Equation~\ref{eq:xi-gate} on \(\Data_t\) before appending the feature value to later contexts.

\begin{runningexamplecontinued}{Testing the missing parent}
The candidate $\xi(p)=\sym{blocked_right}(p)$ divides the mixed group into
\[
\begin{array}{ll}
\xi(p)=\mathrm{false}: & x_t(p)=2\rightarrow x_{t+1}(p)=3,\\
\xi(p)=\mathrm{true}:  & x_t(p)=2\rightarrow x_{t+1}(p)=2.
\end{array}
\]
The gate in Equation~\ref{eq:xi-gate} tests whether this split lowers effect uncertainty and has sufficient support. CEGIS still checks the predicted coordinates.
\end{runningexamplecontinued}

A model-declared fluent $\zeta(o,s)$ is an additional state property proposed by the synthesized model as a Python function of object $o$ and state $s$. It may use one or both arguments and must return a value that can be used as a table key. Returning \code{None} or raising for an object marks the property as inapplicable. The registry accepts at most $N_\zeta$ functions with at most $V_\zeta$ observed values each. For this audit, the outcome signature augments $e$ with horizontal and vertical displacement, rotation and visibility changes, and the pixel-hash transition. An object is engaged when it changes, receives a click, or lies within Chebyshev radius $r_e$ of a changed object. A source action-context cell is mixed when it has at least $n_{\mathrm{mix}}$ samples, at least $B_{\mathrm{mix}}$ outcome signatures, and its most common signature occupies less than fraction $p_{\mathrm{mix}}$.

The engine admits a fluent for a type and cell when it produces at least $B_\zeta$ value bins and each supported bin has at least $n_\zeta$ samples with modal fraction at least $p_\zeta$. Values observed outside engaged samples are reported as requiring further interaction evidence. The registry supplies admitted fluents to the synthesizer as empirically supported state distinctions and refines the actor-facing classes as described in Appendix~\ref{app:eta}.

\subsection{Verifier comparator}
\label{app:verification-details}

State equality in Equation~\ref{eq:replay} is taken through a comparator $V$ that selects which predicted and recorded state fields replay compares. In runs that learn from rendered observations, $V$ compares the predicted and observed palette value at each retained cell. In runs supplied with object records, $V$ compares dynamic objects by identity, position, visibility, rotation, and internal pixels.

Some games display a timer whose rendering changes after each action but whose quantization does not expose a usable state variable. A synthesized model may propose a mask for this display. The verifier accepts the mask only when it is a continuous horizontal or vertical strip with width at most two cells and area at most twice its length. In rendered-observation runs, masked cells are excluded. In object-record runs, an object is excluded only when its display rectangle lies inside the accepted strip. An invalid mask is ignored. This is the qualification on CEGIS in Equation~\ref{eq:replay}.

The loader statically rejects references to the file containing \(\Data\). It permits level-entry caches for reset and level-advance transitions, because those entry states are selected by the environment. The reward function may not read these caches. Transition and reward outputs must also match the recorded successor, reward, and termination flag in \(\Data\).

\subsection{Synthesis, replay, and criticism}
\label{app:synthesis-planning}

The CEGIS procedure \citep{solarlezama2006sketching,jha2010oracle,alur2013sygus} receives the replay buffer \(\Data_t\), preceding Python model, qualitative hypotheses \((\widehat{\mathcal S}_{\mathrm{NL}},\widehat T_{\mathrm{NL}},\widehat R_{\mathrm{NL}})\), and available critic findings. It may revise the extractor, transition and reward functions, planner, and qualitative hypotheses. Candidate programs are evaluated on the accumulated replay buffer; failures provide counterexamples for the next revision. Equation~\ref{eq:replay} defines acceptance on observed transitions.

On scheduled synthesis rounds, an independent critic reviews replay-consistent code for layout-specific constants, omitted interactions, and unsupported generalization. Valid findings prompt repairs followed by another replay check; findings that require new observations are retained as unresolved hypotheses and proposed probes. Replay establishes consistency with the observed transitions, while predictions for unvisited states remain hypotheses.

The synthesis interface specifies object records and declared fluents while leaving code organization flexible. Coupled effects may be implemented jointly across objects; a mechanic need not belong to a single object's method.

For $\widehat s_{\ell+1}=\widehat T(s_\ell,a_\ell)$, the replay mismatch set is
\begin{equation}
  \mathcal M_\ell=
  \{X(\bar o):
    \widehat s_{\ell+1}[X(\bar o)]
    \neq s_{\ell+1}[X(\bar o)]\}.
  \label{eq:mismatch-set}
\end{equation}
Equation~\ref{eq:mismatch-set} identifies the CPF families involved in a replay failure. The static checks described in Section~\ref{app:verification-details} reject replay lookup and reward functions that use cached future states.

\begin{runningexamplecontinued}{Counterexample localization}
If the only mismatch is
\[
  \widehat x_{t+1}(p_1)\ne x_{t+1}(p_1),
\]
then the first repair target is the CPF for $x'(p)$ rather than the complete transition function. The synthesizer can test whether a parent such as $\sym{blocked_right}(p)$ is missing or whether the update body is wrong.
\end{runningexamplecontinued}

\paragraph{Checked planning.} Once \(\Data\) contains reward evidence, the planner searches for a sequence that satisfies
\begin{equation}
  \widehat s_{k+1}=\widehat T(\widehat s_k,a_k),
  \qquad
  \max_{0\leq k<L}
  \widehat R(\widehat s_k,a_k,\widehat s_{k+1})>0.
  \label{eq:model-planning}
\end{equation}
Equation~\ref{eq:model-planning} requires the model rollout to contain positive predicted reward. Before live use, the harness verifies plans from stored initial states of completed levels. During live execution, it compares each observed successor with the model prediction. A disagreement adds a counterexample to \(\Data\) and terminates the remaining plan.  This checked execution is the operational basis of goal-directed falsification: a plan reaches reward or returns evidence against the model that proposed it.

\begin{runningexamplecontinued}{Goal hypothesis}
In the simplified \code{ls20} example, a candidate goal requires an unlocked door and a player at that door:
\[
  G(s)=\mathbf 1\!\left\{\exists p,d:\neg\sym{locked}(d)\wedge\sym{at_door}(p,d)\right\}.
\]
The predicate is retained only when it agrees with the recorded reward transitions, just as a transition CPF must agree with replay.
\end{runningexamplecontinued}

\subsection{Online learning algorithm}
\label{app:agent-loop}

Algorithm~\ref{alg:system-cycle} summarizes the interaction loop. Here \(\mathrm{NL}=(\widehat{\mathcal S}_{\mathrm{NL}},\widehat T_{\mathrm{NL}},\widehat R_{\mathrm{NL}})\), and \textsc{Synthesize} includes the replay and scheduled critic procedure in Appendix~\ref{app:synthesis-planning}. \textsc{RepairGate} uses the batching limits in Appendix~\ref{app:experiments}. While repair is deferred, the actor collects additional evidence; reward-directed programmatic planning requires replay consistency and planner eligibility. Observation abstraction reconstructs any required history from the recorded trajectory, and \textsc{ExploitGoal?} includes the planner eligibility checks.

\begin{algorithm}[H]
\small
\caption{OPINE-World online learning loop (schematic)}
\label{alg:system-cycle}
\begin{algorithmic}[1]
\State \textbf{Input:} known action interface \(\mathcal A\), action budget \(B\)
\State \(\omega\gets\textsc{Env.Reset}();\quad \Data\gets\varnothing\)
\State \(\mathrm{NL},\widehat{\mathcal S}_{\mathrm B},
  \widehat T_{\mathrm P},\widehat R_{\mathrm P},
  \widehat P,\widehat f^{-1}
  \gets\textsc{Initialize}(\omega,\mathcal A)\)
\While{budget remains and the game is unfinished}
  \If{\(\textsc{RepairGate}(\Data)\) and
      \(\neg\Phi_1(\widehat f^{-1},\widehat T_{\mathrm P},
                    \widehat R_{\mathrm P};\Data)\)}
    \Repeat
      \State \(\mathrm{NL},\widehat T_{\mathrm P},\widehat R_{\mathrm P},
        \widehat P,\widehat f^{-1}
        \gets\textsc{Synthesize}(\Data,\mathrm{NL},
                                  \widehat{\mathcal S}_{\mathrm B})\)
    \Until{\(\Phi_1(\widehat f^{-1},\widehat T_{\mathrm P},
                    \widehat R_{\mathrm P};\Data)\)}
  \EndIf
  \State Reconstruct \(\widehat s\) from \(\omega\) and recorded history using \(\widehat f^{-1}\).
  \State \(c\gets\Phi_1(\widehat f^{-1},\widehat T_{\mathrm P},\widehat R_{\mathrm P};\Data)\)
  \If{\(c\) and \(\textsc{ExploitGoal?}(\widehat s,\mathrm{NL},\widehat{\mathcal S}_{\mathrm B})\)}
    \State \(\pi\gets\widehat P(\widehat s;
                    \widehat T_{\mathrm P},\widehat R_{\mathrm P})\)
  \Else
    \State \(\pi\gets\textsc{EpistemicExperiment}(\widehat s,\mathrm{NL},\widehat{\mathcal S}_{\mathrm B})\)
  \EndIf
  \For{\(a\in\textsc{ShortPrefix}(\pi)\), within the remaining action budget}
    \State \((\omega',r)\gets\textsc{Env.Step}(a)\)
    \State \(\Data\gets\Data\cup\{(\omega,a,r,\omega')\}\)
    \State \(\mathrm{NL},\widehat{\mathcal S}_{\mathrm B}
      \gets\textsc{Update}(\Data,\mathrm{NL},
                            \widehat{\mathcal S}_{\mathrm B})\)
    \State \(m\gets\neg\textsc{LiveCheck}_{\Phi_1}
      (\omega,a,r,\omega')\)
    \State \(\omega\gets\omega'\)
    \If{\(m\)}
      \State discard the unexecuted suffix of \(\pi\); \textbf{break}
    \EndIf
  \EndFor
\EndWhile
\end{algorithmic}
\end{algorithm}

\section{Experimental Details}
\label{app:experiments}

The studies use different sources of replication, which makes the experimental unit important for interpreting their uncertainty estimates. Table~\ref{tab:app-configs} summarizes the unit, evidence, and intervention for each study. All language-model roles use Claude Opus 4.8 with a one-million-token context window, except the GPT-5.5 high benchmark run and the MiniGrid comparison, which uses GPT-6 Luna low for both OPINE-World and WorldCoder. The main ablations in RQ1 use high effort, while RQ2, RQ3 ablations use low effort. Runs use Docker filesystem and network isolation to prevent cheating.

The online loop uses $K_{\mathrm{critic}}=3$. After the first prediction failure, model review may be deferred for at most $D_{\mathrm{move}}=5$ executed moves, $D_{\mathrm{error}}=3$ accumulated errors, or $D_{\mathrm{plan}}=3$ completed action plans. These constants are fixed across online conditions. The planner becomes eligible after the first completed level.

\begin{table}[H]
\caption{\textbf{Experimental configurations.} The unit column gives the independent or paired unit used in summaries and tests. Learned extraction infers object records from rendered frames.}
\label{tab:app-configs}
\centering
\scriptsize
\begin{tabular}{@{}L{0.12\linewidth}L{0.18\linewidth}L{0.47\linewidth}@{}}
\toprule
Study & Unit & Evidence and conditions \\
\midrule
RQ1: exploration & 25 paired games & The ablation removes \(\widehat{\mathcal S}_{\mathrm B}\). \\
$\epsilon$-greedy & 25 shared games & Descriptive comparison at a 400-action cutoff. The baseline removes \(\widehat{\mathcal S}_{\mathrm B}\) and explores random objects with probability $\epsilon=0.1$. \\
RQ2: fixed evidence & 5 runs per condition & The conditions receive the same 757-action \code{g50t} trajectory in batches of six. The actor and planner are disabled. Conditions are the full system, monolithic and type-restricted programs, no \(\widehat{\mathcal S}_{\mathrm B}\), and no persistent NL. \\
RQ2: revision & 10 repairs per condition for M1 and M3 & M1 and M3 start from independently synthesized source programs; within each comparison, the 250-action repair trajectory is matched across conditions. M2 measures six level transitions within each of the five fixed-evidence runs. \\
RQ3: online & 5 runs per condition & The full system and the no-persistent-language condition learn the first three \code{g50t} levels with a 1,000-action cap. The ablation rejects prose, comments, and docstrings from the persisted world model. \\
RQ3: five games & 1 run per game and condition & Full system, no persistent NL, and no persistent NL with code comments and docstrings kept, on \code{cd82}, \code{cn04}, \code{ft09}, \code{g50t}, and \code{tn36}. Runs stop at 2,000 actions or once the current level can no longer reach 5\% RHAE. \\
MiniGrid/Sokoban & 6 runs per system & WorldCoder's curriculum and episode protocol, with WorldCoder's text state, change descriptions, action names, and mission given to both systems. \\
\bottomrule
\end{tabular}
\end{table}

\subsection{Scope and interpretation of the ablations}
\label{app:ablation-scope}

\paragraph{Exploration as a capability of the whole system.}
In an exploration task, a working planner enables the agent to plan efficient routes to informative states, programmatic replay verifies that its understanding of the world is consistent with the observed history, object extraction allows principled tracking of the world state, plan soundness determines exploitation or exploration priorities. The capability of the learned engine therefore can be classified as an exploration capability. This is why RQ1 evaluates the contribution of \(\widehat{\mathcal S}_{\mathrm B}\) through this synthesis, verification, planning, and action-selection loop in tandem.

\paragraph{Limiting Controls.}
For any experiment, it is possible to generate an almost unlimited list of potential confounders. When designing these experiments, the issue we faced is not whether a variable exists, but whether it is sufficiently important and plausible to threaten a conclusion. In RQ1, improvements to the learned engine and the resulting plans are mechanisms through which Bayesian guidance can improve exploration. In RQ2, different interaction evidence could itself explain a difference in model quality, so the trajectory is held fixed across conditions. 
\label{app:g50t-motivation}

To assess the world-modeling capabilities targeted by RQ2 and RQ3, we evaluate seven recurring challenges: (1) partial observability, (2) history-conditioned state and dynamics, (3) autonomously moving agents, (4) coupled interactions among multiple objects, (5) context-dependent behavior, (6) intermediate states between actions, and (7) the introduction of new objects and mechanics as tasks progress.

RQ2 and RQ3 use \code{g50t} as a complete case study of these world-modeling challenges. These properties make \code{g50t} a demanding environment for testing both executable model structure and persistent qualitative hypotheses. These modeling requirements generalize beyond this single domain's isolated visual layout. In addition, the large discovery failure under the Bayesian ablation is informative for examining mechanisms behind the benchmark gains. Each of these runs is prohibitively expensive, and the traces we analyzed under this general case resulted in convincing enough behaviors to motivate this approach. For diversity in our RQ3 comparison, we additionally pick 4 random games, of which contain games where the \(\widehat{\mathcal S}_{\mathrm B}\) ablations perform comparably.

\subsection{Statistical analysis}

The tests follow the unit that generated the variation. Actions, levels, and synthesis rounds within a run are not treated as independent observations. All reported tests are two-sided, and the reported $p$-values are unadjusted.

RQ1 pairs conditions within each game. Per-game RHAE uses the Wilcoxon signed-rank test. Level completion first reduces each game to its paired difference in completed levels and then applies a sign-flip randomization test. The paired cluster bootstrap resamples 25 games with replacement 100,000 times. These tests compare the high-effort full and ablated conditions.

The fixed-evidence and M1 analyses match conditions by run index. Their exact sign-flip tests enumerate the $2^n$ signs of the paired differences and count assignments whose absolute mean is at least the observed value. M3 and the online RQ3 study use exact two-sample permutation tests over the independent synthesis replications. These tests enumerate assignments of the pooled observations to conditions while preserving the number of runs per condition. M2 is summarized at the level-transition unit, with transitions grouped within their evolving run.

\subsection{Held-out prediction and program classification}
\label{app:structure-metrics}

Replay accuracy on synthesis evidence cannot measure generalization. The structure study therefore uses a second buffer that is unavailable during synthesis. A seeded random policy generates randomly seeded 100,000 trajectories. This trajectory set is shared by all conditions. The evaluator replays each trajectory, and supplies the learned engine with the ground-truth before-state at each step. Exact match requires its predicted successor to match all object records in the ground-truth successor. This can be called a test set.

A prompt restriction also does not guarantee that synthesis produces the intended source organization. For function $k$, let $c_k$ be cyclomatic complexity, $\ell_k$ be source lines, $d_k$ be decision sites, and $h_k$ be maximum nesting depth. Let $c_{(k)}$ denote complexities sorted from largest to smallest. We classify a program $P$ with
\begin{equation}
\begin{aligned}
C_{\max}(P)&=\max_k c_k,
&C_{95}(P)&=\operatorname{P95}_{\ell_k}(c_k),\\
H_d(P)&=\sum_k\left(\frac{d_k}{\sum_j d_j}\right)^2,
&D_{\max}(P)&=\max_k h_k,\\
C_5(P)&=\sum_{k=1}^{5}c_{(k)}.&&
\end{aligned}
\label{eq:monolithicity-classification}
\end{equation}
Radon 6.0.1 computes per-function source lines and McCabe cyclomatic complexity, which counts independent control-flow paths \citep{mccabe1976complexity,lacchia2023radon}. $C_{\max}$ identifies the largest function, while $C_{95}$ is the source-line-weighted 95th percentile and $C_5$ sums the five largest function complexities. $H_d$ applies the Herfindahl--Hirschman index to each function's share of decision sites \citep{hirschman1964paternity}. It approaches one when the same function contains most decisions. $D_{\max}$ is the largest Python block-nesting depth computed from the abstract syntax tree. Together, these measurements classify monolithicity. Held-out exact match and revision cost assess model outcomes. We also archive maximum cognitive complexity as an auxiliary source-complexity metric \citep{campbell2018cognitive}.

Holding the trajectory fixed and disabling the actor and planner in RQ2 isolates the influence of $\widehat{\mathcal S}_{\mathrm B}$ through synthesis, including the induced program structure and learned transition rules. Online, the synthesized engine also supports action selection, so exploration outcomes reflect the coupled synthesis and exploration loop. RQ2 evaluates synthesis separately under controlled evidence.

The reported diagnostic uses $\alpha_0=1$, which gives each effect one symmetric pseudocount. We set $m_{\min}=2$ so that normalized entropy is defined before two effects have been observed. The base and engagement neighborhoods use $r_c=r_e=1$ because immediate contact captures the local interactions used by the actor while limiting group fragmentation.

The prompt budget admits $K_\xi=3$ high-weight groups, and $\mathcal R_\xi=\{2,3\}$ tests nearby relational context beyond the base neighborhood. The observable-refinement gate uses $\delta_\xi=0.05$, $n_\xi=3$, and $p_\xi=0.95$. The declared-fluent registry uses $N_\zeta=16$, $V_\zeta=12$, $n_{\mathrm{mix}}=3$, $B_{\mathrm{mix}}=2$, $p_{\mathrm{mix}}=0.9$, $B_\zeta=2$, $n_\zeta=2$, and $p_\zeta=0.9$. The capacity limits bound prompt growth. The support and concentration thresholds require repeated evidence before a state distinction is admitted. Equation~\ref{eq:xi-gate} and Section~\ref{app:refinement} define how these constants are used.

\section{Extended Related Work}
\label{app:related-work}

\paragraph{Learning World Models for Planning.}
Planning in unfamiliar environments requires learning action
consequences from limited interaction, while the relevant state
variables and goals may also be unknown.
Latent methods learn compressed dynamics for rollouts and
control \citep{ha2018worldmodels,hafner2025dreamerv3}.
Programmatic methods synthesize executable dynamics for
prediction and search \citep{tang2024worldcoder,ahmed2025theorycoder,poeworld2025}.
Language-augmented methods retain NL theories that
evolve with observed transitions \citep{levy2025worldllm}, and
NL-PDDL plans over a supplied NL action model via lifted
regression with LLM entailment \citep{liu2026natural}, but does not
learn that model from interaction.
Yet fitting observed transitions does not establish whether the
representation captures distinctions needed for untried actions.
Existing approaches also do not unify NL and
programmatic models in a way that provides an actionable measure of
uncertainty of the learned representation. Additionally, planning over high branching factors becomes computationally prohibitive at long horizons with a general planner, motivating task-specific planners and heuristics.

\paragraph{Exploration and State Discovery.}
Discovering these omissions requires interactions that distinguish
competing explanations of action effects.
Information-based methods target uncertainty about model parameters
\citep{lindley1956measure,houthooft2016vime}.
Surprise-based methods use prediction error or disagreement among
transition models \citep{pathak2017curiosity,sekar2020planning}.
State-discovery methods add history features to resolve perceptual
aliasing \citep{mccallum1996reinforcement} or learn dependencies
among supplied variables \citep{strehl2007efficient}.
Yet uncertainty measured within a fixed representation does not by
itself identify which missing object property or relation would
explain conflicting outcomes. Open-world exploration must therefore
also generate and test hypotheses about the representation itself.

\paragraph{ARC-AGI-3 Systems.}
Recent ARC-AGI-3 systems demonstrate the value of persistent
test-time artifacts and structured interaction. \citet{rudakov2025graph}
heuristically prioritize and enumerate
untested frame--action pairs, navigating through recorded transitions
to previously observed frontier states. This improves coverage over
random exploration but does not abstract equivalent states or
prioritize experiments according to their expected information about
the underlying mechanics. Its exploration cost consequently scales
with the reachable visual state space and depends on frontier states
remaining recoverable. Other systems use replay-tested executable
models \citep{rodionov2026executableworldmodelsarcagi3}, multi-agent
workspaces \citep{sarafian2026workspace}, and continual programmatic
harnesses \citep{karten2026continual,furgale2026nooa}. These approaches
provide increasingly capable mechanisms for retaining experience,
constructing models, and coordinating test-time reasoning, but lack formal approaches to efficient, information-informed exploration.

\paragraph{Our Point of Departure.}
We introduce a hybrid NL, Bayesian, and programmatic
world model that unifies representation revision with targeted
exploration. An object-oriented NL proto-representation
preserves partial hypotheses, while a Bayesian belief state prioritizes
sparsely tested or conflicting object--action effects. The resulting
observations support revision of both the proto-representation and
executable model, which supplies predictions and reward-directed
planning.

\section{Limitations}
\label{app:limitations}

\paragraph{Domain assumptions.}
ARC-AGI-3 environments are deterministic and use discrete state and
action spaces. Accordingly, our Bayesian diagnostic interprets mixed
effects as evidence that the learned ontology omits a relevant property,
relation, or historical variable. In stochastic environments, mixed
effects may instead reflect irreducible transition uncertainty.
Extending OPINE-World to stochastic or continuous-time, state, or action
spaces would therefore require distributional transition predictions,
probabilistic replay consistency, and uncertainty estimates. This is beyond the scope of our system.

\paragraph{Unvisited states.}
OPINE-World relies on replay consistency and observed effects; therefore, \(\Phi_1\) gives no guarantee for unvisited states. While our systematic hypothesis generation and goal-conditioned experiment design could prioritize untested objects that may be necessary for reward, a well-principled system for this failure mode based on goal-conditioned relevance, coverage, and structured hypothesis generation would allow better insights into unexplored states.

\paragraph{Supplied symbolic domains.}
\label{app:worldcoder-limits}
OPINE-World does not improve on domain specific approaches, such as WorldCoder, in every case (Table~\ref{tab:minigrid}). In these domains, the symbolic state, action names, and mission may be supplied, which removes the uncertainty that OPINE-World's object abstraction, goal inference, and exploration are designed for. Outside of these domains, WorldCoder is unable to progress as it lacks the systems to acquire this information. Within these domains, due to the rigid heuristic and sub-optimality of its MCTS planner, WorldCoder is unable to progress in certain levels due to intractability (Table~\ref{tab:minigrid}. These are the cases where OPINE-World demonstrate a higher capability.

\paragraph{Computational cost and evaluation coverage.}
A complete 25-game evaluation uses Claude Opus 4.8 at high reasoning effort for acting and program synthesis. Including replay and planning, individual environments can consume up to 2 billion cumulative tokens. Due to this cost, we allocated the available compute to maximize benchmark coverage in areas that are pivotal for our claims. Due to the heavy compute requirements across our ablations, we limited our ablations to Opus 4.8 low, but do not see a barrier to generalization across families.

\section{Additional Results and Artifact Audits}
\label{app:additional-observations}

The paired game audit links exploration failures to the mechanics left unresolved. The fixed-evidence and revision studies compare how the resulting programs predict and change, while the NL ablation tests the value of retaining explanations between synthesis sessions. Examples are grouped by the finding they support, with the measurements and source excerpts reported together.

\subsection{Complete-system benchmark comparison}
\label{app:benchmark-results}

OPINE-World's aggregate advantage includes completing games where the strongest comparator entirely fails (Table~\ref{tab:external-full}). The comparison uses matching game builds and scoring rules, including the baseline1 scorecard rather than the earlier run reported in its paper \citep{rodionov2026executableworldmodelsarcagi3,sarafian2026workspace,karten2026continual,arcscorecards2026}. The per-game results and coverage counts show why both completion and action efficiency matter when interpreting the aggregate score.

\begin{table}[H]
\caption{\textbf{Per-game RHAE and aggregate coverage.} OPINE-World, the \(\widehat{\mathcal S}_{\mathrm B}\) ablation, and other systems. All games run on high, showing standard errors. NOOA reports only aggregate scores.}
\label{tab:external-full}
\centering
\scriptsize
\begin{tabular}{@{}lrrrrrrr@{}}
\toprule
& \multicolumn{3}{c}{OPINE-World} & \multicolumn{4}{c}{Other systems (GPT-5.5)} \\
\cmidrule(lr){2-4}\cmidrule(l){5-8}
Game & \shortstack{GPT-5.5} & \shortstack{Opus 4.8} & \shortstack{No \(\widehat{\mathcal S}_{\mathrm B}\) (Opus 4.8)} & baseline1 & DreamTeam & NOOA & \shortstack{Continual} \\
\midrule
\code{ar25} & \textbf{100.0} & \textbf{100.0} & \textbf{100.0} & \textbf{100.0} & \textbf{100.0} & -- & 86.5 \\
\code{bp35} & \textbf{37.2} & 2.6 & 2.0 & 17.3 & 0.8 & -- & 0.4 \\
\code{cd82} & \textbf{100.0} & \textbf{100.0} & \textbf{100.0} & \textbf{100.0} & 79.0 & -- & 6.3 \\
\code{cn04} & \textbf{100.0} & \textbf{100.0} & 47.6 & \textbf{100.0} & 0.0 & -- & 0.3 \\
\code{dc22} & \textbf{100.0} & 82.4 & 14.3 & 36.4 & 0.0 & -- & 33.7 \\
\code{ft09} & \textbf{100.0} & \textbf{100.0} & \textbf{100.0} & 95.1 & \textbf{100.0} & -- & 88.8 \\
\code{g50t} & 70.8 & 61.7 & 0.0 & \textbf{100.0} & 2.9 & -- & 13.4 \\
\code{ka59} & \textbf{100.0} & 62.5 & 60.6 & 63.0 & 59.7 & -- & 2.3 \\
\code{lf52} & \textbf{61.1} & 4.2 & 3.5 & 34.9 & 10.9 & -- & 2.1 \\
\code{lp85} & \textbf{100.0} & \textbf{100.0} & \textbf{100.0} & \textbf{100.0} & \textbf{100.0} & -- & 43.4 \\
\code{ls20} & 86.0 & 71.2 & 59.1 & \textbf{98.3} & 14.7 & -- & 0.0 \\
\code{m0r0} & \textbf{100.0} & \textbf{100.0} & \textbf{100.0} & 71.6 & 47.6 & -- & 76.7 \\
\code{r11l} & \textbf{100.0} & \textbf{100.0} & \textbf{100.0} & 74.3 & 47.6 & -- & 5.5 \\
\code{re86} & 81.4 & \textbf{100.0} & \textbf{100.0} & 26.4 & 58.3 & -- & 30.3 \\
\code{s5i5} & \textbf{83.1} & 27.3 & 7.5 & 0.7 & 38.2 & -- & 4.8 \\
\code{sb26} & \textbf{100.0} & 89.8 & \textbf{100.0} & \textbf{100.0} & 94.0 & -- & 0.1 \\
\code{sc25} & \textbf{85.0} & 84.0 & 28.6 & 20.1 & 36.1 & -- & 12.8 \\
\code{sk48} & \textbf{56.5} & 21.3 & 16.7 & 30.2 & 6.9 & -- & 6.2 \\
\code{sp80} & 99.3 & \textbf{100.0} & 66.0 & 5.5 & 0.1 & -- & 22.9 \\
\code{su15} & \textbf{92.5} & 92.1 & 50.6 & 88.7 & 2.2 & -- & 3.5 \\
\code{tn36} & \textbf{100.0} & 68.4 & 3.6 & 26.2 & 5.0 & -- & 2.2 \\
\code{tr87} & \textbf{100.0} & \textbf{100.0} & \textbf{100.0} & 86.3 & 12.6 & -- & 81.3 \\
\code{tu93} & 87.9 & \textbf{100.0} & \textbf{100.0} & \textbf{100.0} & \textbf{100.0} & -- & 20.2 \\
\code{vc33} & \textbf{100.0} & 92.4 & 88.9 & 19.4 & 21.4 & -- & 5.4 \\
\code{wa30} & 66.0 & \textbf{100.0} & 5.0 & \textbf{100.0} & 13.3 & -- & 6.2 \\
\midrule
Mean \(\pm\) SE & \textbf{84.3 \(\pm\) 3.5} & 78.4 \(\pm\) 6.3 & 58.2 \(\pm\) 8.3 & 63.7 \(\pm\) 7.4 & 38.1 \(\pm\) 7.6 & 50.2$^{\dagger}$ & 22.2 \(\pm\) 5.9 \\
Levels completed & \textbf{176/183} & 160/183 & 126/183 & 146/183 & 98/183 & 118/183 & 86/183 \\
Games completed & \textbf{21/25} & 20/25 & 12/25 & 14/25 & 6/25 & -- & 5/25 \\
\bottomrule
\end{tabular}
\end{table}

\subsection{$\epsilon$-greedy baseline test}
\label{app:epsilon-400}

\begin{figure}[p]
\centering
\includegraphics[width=\linewidth]{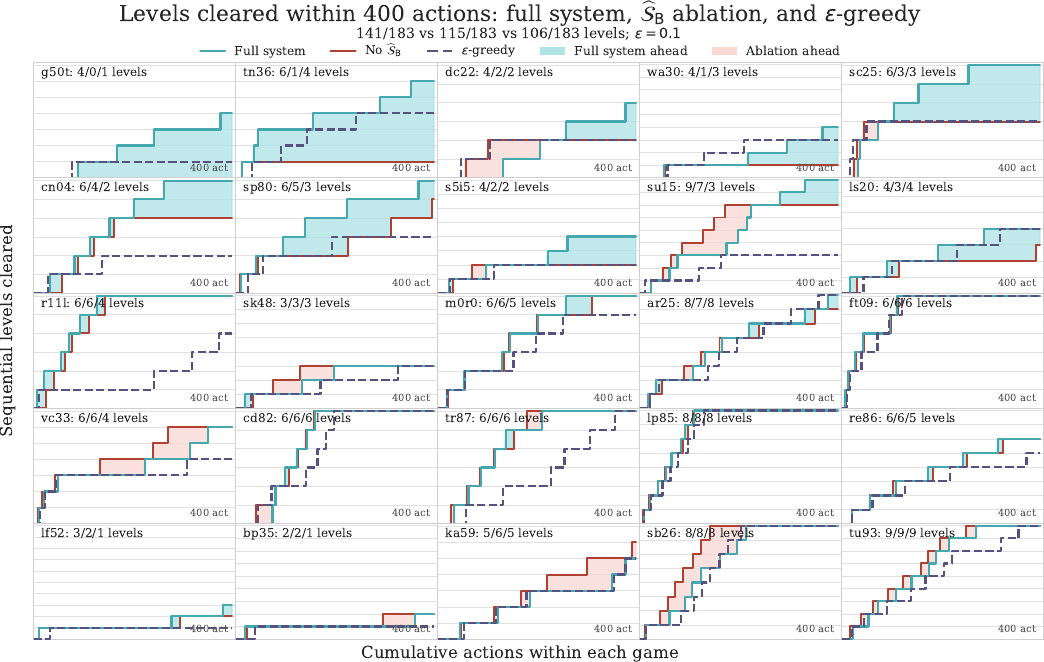}
\caption{\textbf{Levels completed within 400 actions} for the full system, the \(\widehat{\mathcal S}_{\mathrm B}\) ablation, and $\epsilon$-greedy ablation.}
\label{fig:rq1-epsilon-400}
\end{figure}

The $\epsilon$-greedy baseline ablates the \(\widehat{\mathcal S}_{\mathrm B}\) exploration layer and replaces each newly proposed action sequence with probability $\epsilon=0.1$ of blind exploration towards a random object. We pick a 400-action budget cutoff for all comparisons due to budget constraints. The results show a clear conclusion even at this limit (Table~\ref{tab:rq1-epsilon-400} and Figure~\ref{fig:rq1-epsilon-400}). Random exploration is insufficient to match the full system or even the \(\widehat{\mathcal S}_{\mathrm B}\) ablation.

\begin{table}[H]
\caption{\textbf{Exploration at 400 actions per game.} RHAE follows Appendix~\ref{app:scoring}.}
\label{tab:rq1-epsilon-400}
\centering
\small
\begin{tabular}{@{}lrrr@{}}
\toprule
Condition & RHAE & Levels cleared & Games solved \\
\midrule
Full system & 65.97 & 141/183 & 13/25 \\
\(\widehat{\mathcal S}_{\mathrm B}\) ablated & 52.96 & 115/183 & 8/25 \\
$\epsilon$-greedy ($\epsilon=0.1$) & 39.49 & 106/183 & 7/25 \\
\bottomrule
\end{tabular}
\end{table}

\subsection{Paired level progression}
\label{app:level-progression}
\begin{table}[H]
\caption{\textbf{Paired online comparison for RQ1.} Conditions are paired within each of the 25 games.}
\label{tab:rq1-statistics}
\centering
\small
\begin{tabular}{@{}lr@{}}
\toprule
Measurement & Value \\
\midrule
Full-system mean RHAE & 78.40 \\
Ablated mean RHAE & 58.16 \\
Mean paired RHAE difference & 20.24 \\
Standard deviation of paired differences & 29.45 \\
Paired standardized effect ($d_z$) & 0.69 \\
RHAE: exact $p$ & $4.27\times10^{-4}$ \\
\addlinespace
Full-system levels completed & 160/183 \\
Ablated levels completed & 126/183 \\
Additional levels completed & 34 \\
Completion-rate difference (percentage points) & 18.6 \\
Paired game-cluster bootstrap 95\% CI (percentage points) & $[8.0,31.0]$ \\
Level completion: exact $p$ & $1.95\times10^{-3}$ \\
\bottomrule
\end{tabular}
\end{table}

A higher score can reflect either shorter solutions or access to later levels. The full system completes more levels than the \(\widehat{\mathcal S}_{\mathrm B}\) epistemic ablation, with the difference extending beyond the early levels (Table~\ref{tab:level-progression}). Bayesian guidance therefore improves coverage as well as the aggregate score. Its benefit cannot be explained solely by taking fewer actions on levels that both conditions solve.
\begin{table}[H]
\caption{\textbf{Paired progression by level index.} RQ1 completion counts for all 25 games. Eligible gives the number of games containing that level. The final columns count discordant condition pairs. Level positions are sequentially dependent within a game and are not treated as independent samples.}
\label{tab:level-progression}
\centering
\small
\begin{tabular}{@{}rrrrrr@{}}
\toprule
Level & Eligible & Full & No \(\widehat{\mathcal S}_{\mathrm B}\) & Full only & Ablation only \\
\midrule
1  & 25 & 25 & 24 & 1 & 0 \\
2  & 25 & 25 & 23 & 2 & 0 \\
3  & 25 & 24 & 20 & 4 & 0 \\
4  & 25 & 23 & 16 & 7 & 0 \\
5  & 25 & 21 & 15 & 6 & 0 \\
6  & 25 & 21 & 14 & 7 & 0 \\
7  & 16 & 11 & 8  & 3 & 0 \\
8  & 11 & 7  & 5  & 2 & 0 \\
9  & 5  & 3  & 1  & 2 & 0 \\
10 & 1  & 0  & 0  & 0 & 0 \\
\midrule
Positions & 183 & 160 & 126 & 34 & 0 \\
\bottomrule
\end{tabular}
\end{table}

The progression analysis compares completed-level counts within games because later levels depend on earlier completion. The significance test uses these paired game differences, and the confidence interval resamples games while preserving their level sequences (Table~\ref{tab:rq1-statistics}).

\subsection{How online exploration failures arose}
\label{app:exploration-failures}

The largest losses under the epistemic ablation occur when the agent fails to discover a required mechanic. It may continue rearranging familiar objects or repeating similar actions without discovering what prevents completion. This low-diversity exploration can be catastrophic because additional combinations of known actions do not reveal the missing interaction. The paired audit identifies where Bayesian guidance changes this behavior by directing attention toward sparsely modeled interactions (Table~\ref{tab:exploration-failures}).

The audit examines action traces, hypotheses, synthesis handoffs, and observed frame changes across the paired runs. The failure categories distinguish repeated search over familiar mechanics, overlooked interactions, and failures to transfer previously learned behavior.
\begin{table}[H]
\caption{\textbf{Hand-checked exploration behavior.} Patterns found in the online ablation artifacts. The final column summarizes the evidence recorded in the corresponding interaction traces.}
\label{tab:exploration-failures}
\centering
\small
\begin{tabular}{@{}L{0.18\linewidth}L{0.16\linewidth}L{0.58\linewidth}@{}}
\toprule
Pattern & Games & Archived observation \\
\midrule
Repeated or narrow search & \code{su15}, \code{g50t}, \code{dc22}, \code{sk48} &
The actor repeats similar action sequences or block configurations while
unresolved objects and interactions remain untested. \\
Attention remains on familiar objects & \code{wa30}, \code{ar25}, \code{m0r0} &
The actor continues to probe a small set of known objects or mechanics after
they stop producing progress. \\
Untested or misread mechanic & \code{cn04}, \code{sc25}, \code{tn36} &
The actor overlooks essential blatantly presented frame information unimpeded. \\
Missing class-level record & \code{s5i5}, \code{ls20} &
The actor does not use previous tests of the same object class to select a
different interaction or transfer a behavioral property. \\
Incorrect carry-forward model & \code{sp80} &
Limited probing in earlier levels leaves the transition program and planner
with an incorrect mechanic in a later level. \\
\bottomrule
\end{tabular}
\end{table}

\begin{lstlisting}[basicstyle=\ttfamily\scriptsize,caption={Example 1. Failed docking arrangements in the epistemic ablation of \code{cn04}.}]
CONFIGS TRIED, ALL reward=0/done=False:
2. foot-on-body assembly w/ body-overlap: placed (messy pile), NO reward.
3. 7==7 GROW overlay: grew YELLOW to 7 feet, moved G1/G2/G3 so all 7 target feet coincide EXACTLY with YELLOW 7 feet (verified pixel-exact) -> gray3 all, NO reward.
\end{lstlisting}

\begin{lstlisting}[basicstyle=\ttfamily\scriptsize,caption={Example 2. Further docking hypotheses in the same world model.}]
CALL66: 7==7 overlay WITH YELLOW ACTIVE at end -> val3=63 (all 7 YELLOW feet gray3 docked), NO REWARD.
Only untried class: CLEAN rotation-only NON-OVERLAPPING interlock (like L4 spirals).
\end{lstlisting}

\begin{lstlisting}[basicstyle=\ttfamily\scriptsize,caption={Example 3. Later actor notes returning to the docking construction.}]
Step 928
Yellow-docker approach FALSIFIED: reached feet==all-7-comb-tips exactly (via click and via translate) with reward=0.
Step 929
Fresh board, yellow active grow0(3 feet).
Step 972
All 3 combs docked onto yellow 7 feet.
\end{lstlisting}

These traces show combinatorial exploration of familiar docking arrangements while the interaction required for reward remains unresolved. The agent varies the construction without finding a new mechanic, which can be catastrophic for completion when the existing hypothesis is wrong. This is the low-diversity behavior identified in the paired exploration audit. Bayesian guidance addresses it by directing attention toward sparsely modeled interactions instead of relying on further combinations of familiar actions.

The audit also includes cases where Bayesian guidance provides no clear benefit or introduces a cost. Seven pairs show no meaningful difference in the mechanics tested. In \code{tr87}, the agent investigates uncertainty unrelated to the goal, while the \code{sb26} review attributes the ablation's advantage to a planner error following a context reset in the full run. Three pairs remain unclassified because their review notes do not give a reason. These cases distinguish useful discovery from additional exploration that consumes budget without resolving a requirement for progress.

A decrease in ontology error does not necessarily mean that the agent is closer to solving the level. Revised extraction or type assignments can regroup earlier observations and raise $\eta$, while a new level can introduce poorly sampled interactions. The uncertainty traces describe the model's unresolved effects. Progress depends on whether those effects matter for the goal. Traces for \code{ar25}, \code{tr87}, and \code{vc33} begin after extraction stabilizes and describe that later part of learning.

\subsection{Hypothesis-and-test episodes}
\label{app:hypothesis-traces}

The full-system traces show how unresolved mechanics manifest as experiments. The agent spends a sequence of actions exploring a hypothesized solution requirement that is unmodeled by the current programmatic model, incorporates the resulting observation, and uses the revised mechanic in a subsequent plan.

\begin{lstlisting}[basicstyle=\ttfamily\scriptsize,caption={Example 1. \code{sp80} full-system reasoning log.}]
PLAN from step72 [barX inactive@32-51, paddle active@40-63]: [click(40,29) activate barX, 1 (up ->r24-27), 5 fire]. Predict all 3 cups fill.
## CALL23 (step74 fire, PNG-decoded by subagent): ALL 3 CUPS FILLED SIMULTANEOUSLY (color13, ticks18-24) yet REWARD 0.
=> 'all cups filled' is NOT the win condition. HUD row63 = 56 cells left (~84 moves) => budget NOT the issue.
\end{lstlisting}

\begin{lstlisting}[basicstyle=\ttfamily\scriptsize,caption={Example 2. \code{vc33} full-system reasoning log.}]
- ROOT CAUSE: `_l7_band_cross` only ferried ONE marker (left-side preferred),
  clearing it and IGNORING the far-side marker. When BOTH flanking reservoirs
  hold a marker the env SWAPS them: each glyph crosses to the OTHER reservoir,
  re-centred horizontally at its own rows. Because the two glyphs are congruent
  and at the same rows here, the swap reduces to a bar-COLOUR exchange in place
\end{lstlisting}

\begin{lstlisting}[basicstyle=\ttfamily\scriptsize,caption={Example 3. \code{dc22} full-system reasoning log.}]
- CONVEYOR CLICK LOCK (steps 681,690 are NET NO-OPS, before==after): when GREEN
  is riding a navy/teal bar and advancing that bar +2px (or wrapping it from the
  dock) would leave green OFF the bar's new span, the env REFUSES THE WHOLE click
  (every lane frozen) -- the same toggle-locked-under-player rule the L4/L5 navy
  cycles obey. Discrimination vs the FIRING clicks (e.g. step 680: green at c18-19
  on a leftmost-14 bar -> bar slides to span c16-21, green STILL on it -> fires).
\end{lstlisting}

\begin{lstlisting}[basicstyle=\ttfamily\scriptsize,caption={Example 4. \code{sc25} full-system world model.}]
- TWO legends: DIAMOND/15 and MIDDLE-COL/6. RESOLVED -- BOTH are active, each
  with its own reaction: solving diamond/15 SHRINKS the handle (step 83); solving
  middle-col/6 fires a PROJECTILE (step 79 missed because the 4x4 straddled the
  arm). So the player must FIRST shrink (diamond) to fit the rows29-30 arm, THEN
  align + fire (middle-col) down the arm to clear the box+gate, THEN drive RIGHT
  into the dock. The CHECK reaction is keyed by legend color, NOT level.
\end{lstlisting}

These examples show discovery of new behavior rather than repeated combinations of familiar actions. A failed prediction or unsuccessful search leads to a test of the missing mechanic. Bayesian guidance supports this change in attention, while the persistent world model retains the discovered relationship for synthesis and planning. Together, these behaviors explain how exploration can recover progress after the current model stops providing a route to completion.

\subsection{Program organization under fixed evidence}
\label{app:program-audit}

A survey over our synthesized world models find that most programs combine centralized transition logic with object-related helpers. Their updates are commonly organized around a mechanic or level while their state remains object-indexed. This agrees with the RQ2 finding that coupled effects are difficult to assign to independent object rules. \(\widehat{\mathcal S}_{\mathrm B}\) Code Structure allows synthesis to couple objects families in code according to the relationships defined in \(\widehat{\mathcal S}_{\mathrm B}\).

We find that with the same exploration trajectory supplied to synthesis, \(\widehat{\mathcal S}_{\mathrm B}\) Code Structure has the highest mean held-out prediction accuracy. Type-Oriented Code Structure uses more output tokens without recovering that accuracy (Tables~\ref{tab:fixed-outcomes} and~\ref{tab:fixed-synthesis}). A more prescribed organization therefore does not necessarily produce a better predictive model.
\begin{table}[H]
\caption{\textbf{Program-structure classification.} Means and standard deviations over runs measuring code metrics. These verify that the separate arms produce distinct program structures.}
\label{tab:monolithicity}
\centering
\scriptsize
\begin{tabular}{@{}lrrrrr@{}}
\toprule
Condition & $C_{\max}$ & $C_{95}$ & $H_d$ & $D_{\max}$ & $C_5$ \\
\midrule
Full system & $38.2\pm10.2$ & $37.8\pm10.8$ & $.048\pm.009$ & $5.0\pm1.2$ & $117.2\pm16.3$ \\
Monolithic & $257.0\pm82.5$ & $257.0\pm82.5$ & $.476\pm.127$ & $6.6\pm0.9$ & $337.6\pm88.9$ \\
Type-restricted & $49.2\pm35.5$ & $48.9\pm35.6$ & $.056\pm.039$ & $4.4\pm0.5$ & $126.0\pm39.5$ \\
No \(\widehat{\mathcal S}_{\mathrm B}\) & $51.8\pm31.3$ & $51.8\pm31.3$ & $.070\pm.039$ & $5.0\pm0.7$ & $134.0\pm31.1$ \\
No persistent NL & $40.6\pm15.8$ & $40.6\pm15.8$ & $.045\pm.018$ & $4.6\pm0.9$ & $110.0\pm24.7$ \\
\bottomrule
\end{tabular}
\end{table}
\begin{table}[t]
\centering
\caption{\textbf{Program performance and structure over 100,000 randomly sampled trajectories.}}
\label{tab:fixed-outcomes}
\small
\begin{tabular}{@{}lrrr@{}}
\toprule
Condition & Exact match (\%) & SLOC & Functions \\
\midrule
Full system & $83.9\pm9.3$ & $1004\pm84$ & $61.0\pm7.2$ \\
Monolithic & $73.2\pm16.5$ & $915\pm146$ & $18.2\pm4.3$ \\
Type-restricted & $64.4\pm23.6$ & $1107\pm227$ & $71.4\pm7.7$ \\
No $\widehat{\mathcal S}_{\mathrm B}$ & $71.6\pm11.9$ & $949\pm124$ & $48.4\pm4.9$ \\
No persistent NL & $66.9\pm18.5$ & $709\pm61$ & $70.6\pm7.6$ \\
\bottomrule
\end{tabular}
\end{table}
\begin{table}[H]
\centering
\caption{\textbf{Synthesis metrics in RQ2 fixed-trajectory runs.}}
\label{tab:fixed-synthesis}
\centering
\scriptsize
\begin{tabular}{@{}lrrrr@{}}
\toprule
Condition & Rounds & Output (k) & Input (M) & Total (M) \\
\midrule
Full system & $33.6\pm5.7$ & $741\pm184$ & $66.8\pm16.6$ & $67.5\pm16.8$ \\
Monolithic & $29.0\pm3.9$ & $631\pm161$ & $53.2\pm15.3$ & $53.9\pm15.4$ \\
Type-restricted & $34.8\pm9.9$ & $814\pm272$ & $70.7\pm22.7$ & $71.5\pm22.9$ \\
No \(\widehat{\mathcal S}_{\mathrm B}\) & $33.8\pm5.0$ & $789\pm144$ & $65.5\pm10.6$ & $66.3\pm10.7$ \\
No persistent NL & $32.0\pm4.2$ & $722\pm193$ & $64.4\pm15.2$ & $65.2\pm15.3$ \\
\bottomrule
\end{tabular}
\end{table}

\subsubsection{Qualitative Insights}
\label{app:engine-audit}

\begin{lstlisting}[basicstyle=\ttfamily\scriptsize,caption={Example 1. History retained by \(\widehat{\mathcal S}_{\mathrm B}\) Code Structure.}]
if _sig(state) != _ACC["last_sig"]:
        # State does not continue our last output (fresh level / planner branch
        # / reset). Drop stale history; freeze any ghosts present (unknown tapes).
        _ACC["tape"] = []
        _ACC["ghosts"] = [
            {"tape": [], "idx": 0, "x": g["x"], "y": g["y"]}
            for g in state if g.get("name") == PLAYER and not _has9(g)
        ]
\end{lstlisting}

\begin{lstlisting}[basicstyle=\ttfamily\scriptsize,caption={Example 2. History retained by No Natural Language.}]
sig_in = _sig(state)
    if _RUNTIME["sig"] is not None and sig_in == _RUNTIME["sig"]:
        rt = {"n": _RUNTIME["n"], "hist": list(_RUNTIME["hist"]),
              "deposits": [dict(d) for d in _RUNTIME["deposits"]],
              "gidx": _RUNTIME["gidx"], "v": _RUNTIME.get("v", (STEP, STEP)),
              "maxis": _RUNTIME.get("maxis")}
    else:
        rt = {"n": 0, "hist": [], "deposits": [], "gidx": 0, "v": (STEP, STEP), "maxis": None}
\end{lstlisting}

\begin{lstlisting}[basicstyle=\ttfamily\scriptsize,caption={Example 3. The unablated world model records why the gate explanation changed.}]
## RUN-24: latching (co-planar) gate is a TOGGLE
Co-planar gate `qinemeqqdw` oscillates between home and home+slide on each plate
RISING edge (any actor). Falling edge = no move (latched). RECORD/RESET restore
home. Supersedes the "latch-forever" reading (L2/L3 never double-pressed without
a record, so they were consistent with either; L5 s572+s578 double-press forces
the toggle). Momentary above-player gates unchanged. See `gate_ext` accumulator.
\end{lstlisting}

Both the unablated and No NL engines represent movement history in Python. The difference concerns the record of why a rule was adopted and what remains uncertain. The unablated artifacts retain rejected explanations and the observations that replaced them, giving later synthesis access to reasoning that No NL removes.

\begin{lstlisting}[basicstyle=\ttfamily\scriptsize,caption={Shared context before type dispatch in Type-Oriented Code Structure.}]
ctx = _make_ctx(state, action_id)
    new_state = [TYPE_RULES.get(object_type(o), _static_rule)(o, state, action_id, ctx)
                 for o in state]
\end{lstlisting}

Type-Oriented Code Structure still computes shared context before dispatching its object rules. The handlers depend on information about other objects despite their separate organization. This supports the RQ2 argument that object-indexed state can coexist with shared transition logic, which the unablated condition allows synthesis to organize around the interaction.

\subsection{Revision protocol and results}
\label{app:repair-controls}

\begin{figure}[H]
\centering
\includegraphics[width=0.90\linewidth]{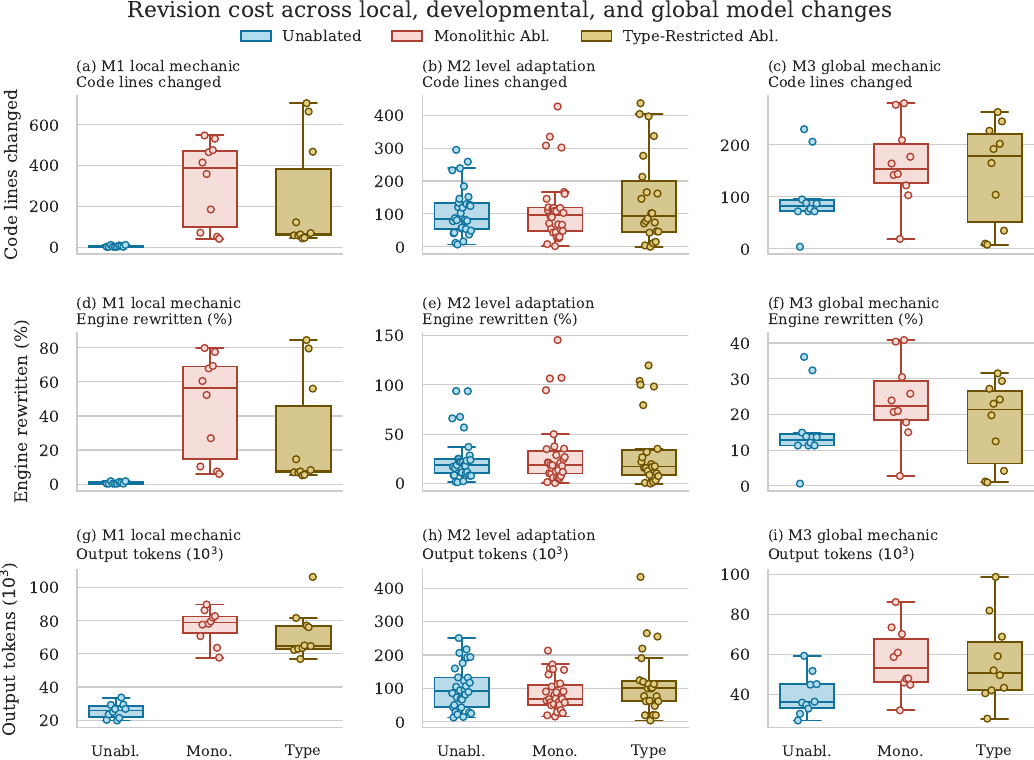}
\caption{\textbf{Full distributions of code and synthesis required by three model revisions.} M1 and M3 show ten repair runs per condition. M2 attributes changes to objects and relations introduced at each level transition in the fixed-trajectory runs. Code lines exclude comments and blank lines. Engine rewrite divides changed code by entering engine size.}
\label{fig:rq2-revisions-full}
\end{figure}

The advantage of \(\widehat{\mathcal S}_{\mathrm B}\) Code Structure is clearest when an existing mechanic changes locally. M1 and M3 compare repair metrics under evidence matched across conditions. M2 measures revisions as later levels introduce mechanics. This separates a local correction from additions that require broader changes to the model (Figure~\ref{fig:rq2-revisions-full}).

\paragraph{Interventions and scope of replication.}
M1 rotates the sliding doors' orientation-to-displacement mapping by $90^\circ$, changing one existing object mechanic. M3 changes the global timer to decrement every third move instead of every second, increasing the available action allowance. Both interventions preserve the level layouts and goal conditions. Repetition across evidence trajectories tests sensitivity to which transitions reveal that modification, while repetition across independently synthesized starting programs tests sensitivity to the model being repaired. These repetitions support conclusions about repair under the tested interventions, including evidence for localized repair in M1. Each of these runs is prohibitively expensive, and the observed benefits under this general case were convincing enough to motivate this approach.
\begin{table}[H]
\caption{\textbf{Detailed revision measurements.} M1 and M3 are means over ten repairs per condition. Code lines exclude comments and blank lines.}
\label{tab:repair-summary}
\centering
\scriptsize
\begin{tabular}{@{}llrrrrr@{}}
\toprule
Change & Condition & Code lines & Rewrite (\%) & Definitions & Rounds & Output (k) \\
\midrule
M1: local & Full system & 4.8 & 0.75 & 1.1 & 2.00 & 25.6 \\
& Monolithic & 314.4 & 45.83 & 9.4 & 5.00 & 76.9 \\
& Type-restricted & 229.6 & 27.50 & 18.2 & 5.00 & 71.7 \\
\addlinespace
M2: level & Full system & 107.5 & 25.92 & 11.8 & 4.70 & 101.5 \\
& Monolithic & 115.7 & 30.57 & 4.1 & 3.93 & 84.5 \\
& Type-restricted & 146.3 & 34.01 & 13.4 & 5.05 & 117.8 \\
\addlinespace
M3: global & Full system & 100.3 & 15.75 & 10.8 & 4.00 & 39.8 \\
& Monolithic & 163.7 & 23.86 & 7.5 & 4.00 & 56.8 \\
& Type-restricted & 145.1 & 17.38 & 12.1 & 4.00 & 56.4 \\
\bottomrule
\end{tabular}
\end{table}

For M1, \(\widehat{\mathcal S}_{\mathrm B}\) Code Structure requires less rewriting, fewer synthesis rounds, and fewer output tokens than the restricted structures (Tables~\ref{tab:repair-summary} and~\ref{tab:repair-tests}). The source examples below show how the same local intervention can lead to either a small correction or a broad revision.
\begin{table}[H]
\caption{\textbf{Tests of revision cost against the full system.} M1 uses exact two-sided sign-flip tests. M3 uses exact two-sample permutation tests. All $p$-values are unadjusted.}
\label{tab:repair-tests}
\centering
\small
\begin{tabular}{@{}llrr@{}}
\toprule
Change & Outcome & Monolithic $p$ & Type-restricted $p$ \\
\midrule
M1 & Changed lines & .00195 & .00195 \\
& Engine churn & .00195 & .00195 \\
& Synthesis rounds & .00195 & .00195 \\
& Output tokens & .00195 & .00195 \\
& Changed definitions & .00586 & .00195 \\
\addlinespace
M3 & Output tokens & .0117 & .0356 \\
& Changed lines & .0700 & .2448 \\
\bottomrule
\end{tabular}
\end{table}

When later levels introduce new mechanics, the revision costs are closer across conditions. The added behavior can require changes across interacting objects even when existing rules have separate helpers. M2 compares the entering engine with the first replay-consistent engine after the new level's evidence, retaining earlier transitions in replay. Its observations are grouped within the evolving programs, and some Type-Oriented Code Structure snapshots are unavailable.

For the global rule change, the conditions take the same number of synthesis rounds, while \(\widehat{\mathcal S}_{\mathrm B}\) Code Structure uses fewer output tokens. The reduction comes from generating less material during repair rather than completing fewer attempts (Table~\ref{tab:repair-tests}).

A small repair is useful when it also incorporates the changed behavior. The M1 prediction check evaluates unseen transitions from the modified world alongside edit size. The unablated model improves after repair, supporting the finding that a local code correction can update the mechanic without a broad rewrite.

\subsection{NL intervention and online results}
\label{app:nl-intervention}

RQ3 removes persistent NL model content while retaining the actor and synthesis roles. Comments and docstrings are removed from the persisted program, while observations, verifier feedback, and communication artifacts remain available. This tests whether retaining hypotheses and explanations helps later synthesis beyond the executable state and evidence available to both conditions. Table~\ref{tab:rq3-games} gives the per-game values behind Figure~\ref{fig:rq3-hybrid}. Its token counts are run totals over all agents and all levels reached, while panels c and e of the figure count synthesis only, on the levels that all three conditions cleared.
\begin{table}[H]
\caption{\textbf{NL ablation across games (Claude Opus 4.8 low).} \emph{NLC} retains code comments and docstrings. Actions, rounds, and tokens are per level over the first \(k\) levels, which every condition cleared.}
\label{tab:rq3-games}
\centering
\scriptsize
\setlength{\tabcolsep}{3pt}
\resizebox{\linewidth}{!}{%
\begin{tabular}{@{}lrcrrrcrrrcrrrcrrrcrrr@{}}
\toprule
& && \multicolumn{3}{c}{\shortstack{RHAE}} && \multicolumn{3}{c}{\shortstack{Levels\\completed}} && \multicolumn{3}{c}{\shortstack{Actions\\per level}} && \multicolumn{3}{c}{\shortstack{Synthesis rounds\\per level}} && \multicolumn{3}{c}{\shortstack{Synthesis tokens\\per level (M)}} \\
\cmidrule(lr){4-6}\cmidrule(lr){8-10}\cmidrule(lr){12-14}\cmidrule(lr){16-18}\cmidrule(l){20-22}
Game & \(k\) && Full & NL & NLC && Full & NL & NLC && Full & NL & NLC && Full & NL & NLC && Full & NL & NLC \\
\midrule
\code{cd82} & 2 &  & \textbf{91.1} & 7.9 & 7.4 &  & \textbf{6} & 2 & 2 &  & 28 & \textbf{27} & 34 &  & 5.0 & \textbf{4.0} & 6.0 &  & \textbf{7.1} & 11.1 & 9.8 \\
\code{cn04} & 1 &  & \textbf{54.8} & 3.1 & 48.0 &  & \textbf{4} & 1 & \textbf{4} &  & \textbf{14} & 36 & 38 &  & \textbf{1.0} & 6.0 & 4.0 &  & \textbf{3.6} & 14.3 & 6.1 \\
\code{ft09} & 4 &  & 90.1 & 45.8 & \textbf{100.0} &  & \textbf{6} & 4 & \textbf{6} &  & 21 & \textbf{16} & \textbf{16} &  & 2.2 & \textbf{1.2} & 1.8 &  & 2.8 & \textbf{2.7} & 3.6 \\
\code{g50t} & 3 &  & \textbf{24.6} & \textbf{24.6} & 20.8 &  & 3 & 3 & 3 &  & \textbf{88} & 117 & 145 &  & \textbf{5.3} & 7.7 & 8.0 &  & \textbf{11.9} & 28.3 & 22.1 \\
\code{tn36} & 2 &  & \textbf{18.2} & 4.7 & 12.3 &  & \textbf{3} & 2 & 2 &  & 78 & 82 & \textbf{36} &  & 4.0 & 3.5 & \textbf{2.0} &  & \textbf{3.3} & 5.0 & 4.1 \\
\midrule
Mean \(\pm\) SE &  &  & \textbf{55.8 \(\pm\) 15.5} & 17.2 \(\pm\) 8.1 & 37.7 \(\pm\) 17.1 &  & \textbf{4.4 \(\pm\) 0.7} & 2.4 \(\pm\) 0.5 & 3.4 \(\pm\) 0.7 &  & \textbf{46.0 \(\pm\) 15.4} & 55.4 \(\pm\) 19.0 & 53.7 \(\pm\) 23.1 &  & \textbf{3.5 \(\pm\) 0.8} & 4.5 \(\pm\) 1.1 & 4.3 \(\pm\) 1.2 &  & \textbf{5.7 \(\pm\) 1.7} & 12.3 \(\pm\) 4.5 & 9.2 \(\pm\) 3.4 \\
\bottomrule
\end{tabular}
}
\end{table}

The same pattern extends to prediction when interaction is held fixed. The unablated models have higher mean accuracy than No NL on unseen trajectories (Table~\ref{tab:fixed-outcomes}). Persistent NL can therefore help synthesis construct stronger world models, in addition to helping the actor choose useful actions.

\subsection{Computation and checked planning}
\label{app:context-accounting}

Repeated synthesis is a major source of computational cost because successive sessions revisit accumulated evidence. The recorded synthesis work totals 174 hours across the full-system benchmark runs. Separate actor and synthesis contexts limit what must be carried into action selection because the actor receives selected model artifacts and evidence summaries rather than the full synthesis transcripts.

Table~\ref{tab:compute} compares score under wall-clock budgets and inference cost. On GPT-5.5 high, OPINE-World costs less than baseline1 on the same model and scores higher at all wall-clock budgets. NOOA, whose authors designed around a two-hour limit, scores higher at that budget (50.2 vs.\ 38.9) and at a far lower cost (\$446 vs.\ \$17,334). We do not claim that OPINE-World is superior under highly limited compute and time constraints. This is not the setting we optimize for; our aim is to demonstrate the value of our exploration mechanism.

\begin{table}[H]
\caption{\textbf{Score under wall-clock budgets and inference cost.} Tokens are input (B) / output (M).}
\label{tab:compute}
\centering
\scriptsize
\setlength{\tabcolsep}{4pt}
\begin{tabular}{@{}llrrrrrrrr@{}}
\toprule
& & \multicolumn{5}{c}{RHAE within wall-clock budget} & & \multicolumn{2}{c}{Cost (USD)} \\
\cmidrule(lr){3-7}\cmidrule(l){9-10}
Method & Model & 2\,h & 3\,h & 4\,h & 8\,h & Final & Tokens (B/M) & 25 games & Per game \\
\midrule
OPINE-World & GPT-5.5 high & 38.9 & \textbf{50.0} & \textbf{58.7} & \textbf{82.9} & \textbf{84.3} & 17.6 / 19.9 & \$17,334 & \$693.36 \\
baseline1 & GPT-5.5 high & 28.2 & 36.8 & 47.2 & 54.4 & 63.7 & 28.6 / 49.3 & \$18,600 & \$744.00 \\
NOOA & GPT-5.5 high & \textbf{50.2}$^{\dagger}$ & -- & -- & -- & 50.2$^{\dagger}$ & -- & \$446$^{\dagger}$ & \$17.85$^{\dagger}$ \\
DreamTeam & GPT-5.5 high & -- & -- & -- & -- & 38.1 & -- & \$18,000 & \$720.00 \\
Continual & GPT-5.5 high & 14.5 & 15.5 & 15.5 & 15.5 & 22.2 & 0.46 / 4.3 & \$977 & \$39.06 \\
\bottomrule
\end{tabular}
\par\smallskip\raggedright\footnotesize $^{\dagger}$ NOOA's paper reports a single cost figure without token or call counts and doesn't expose any artifacts, and as such we couldn't verify whether this cost was measuring pure visible output, or with hidden reasoning included. If only measuring visible output, the true cost of this run may be higher than reported here.
\end{table}

\begin{lstlisting}[basicstyle=\ttfamily\scriptsize,caption={Example 1. Token-position mappings used by the \code{lp85} planner.}]
for op in ops:
        mp = {}
        for keys in op["cycles"]:
            ids = [idx[k] for k in keys]
            n = len(ids)
            for k in range(n):
                src = ids[(k + 1) % n] if op["ccw"] else ids[(k - 1) % n]
                mp[src] = ids[k]
        perms.append((mp, op["click"]))
\end{lstlisting}

\begin{lstlisting}[basicstyle=\ttfamily\scriptsize,caption={Example 2. Transition and reward checks in the \code{ar25} planner.}]
nxt = transition_function(cur, a)
            r, done = reward_function(cur, a, nxt)
            new_path = path + [a]
            if r and r > 0 and done:
                return new_path
\end{lstlisting}

The learned program also gives the planner a smaller representation of the task. In \code{lp85}, button presses become permutations of token positions. The planner can search over these positions instead of rendered frames. Candidate solutions remain dependent on the learned transition and reward rules, which are checked during execution. This combination makes the program useful for finding actions as well as predicting them.


\section{Scoring and Protocol Checks}
\label{app:scoring}
\label{app:protocol}

For game $g$, let $L_g$ be its number of levels, let $c_g$ be the number completed in order, and let $b_{g\ell}$ and $a_{g\ell}$ be the human and agent action counts for completed level $\ell$. The level contribution and total level weight are
\begin{equation}
  z_{g\ell}=\begin{cases}
    \min\!\left\{(b_{g\ell}/a_{g\ell})^2,1.15\right\},
      &\ell\le c_g,\\
    0,&\ell>c_g,
  \end{cases}
  \qquad
  W_g=\sum_{\ell=1}^{L_g}\ell.
  \label{eq:rhae-level}
\end{equation}
Equation~\ref{eq:rhae-level} assigns zero to incomplete levels while retaining their weight in $W_g$. The completed-weight cap prevents an early bonus from raising a game above the fraction of level weight completed. For a set of $G$ games,
\begin{equation}
\begin{aligned}
  S_g&=100\min\!\left\{
    \frac{\sum_{\ell=1}^{L_g}\ell z_{g\ell}}{W_g},
    \frac{\sum_{\ell=1}^{c_g}\ell}{W_g}\right\},\\
  \mathrm{RHAE}&=\frac{1}{G}\sum_{g=1}^{G}S_g.
  \label{eq:rhae}
\end{aligned}
\end{equation}
Equation~\ref{eq:rhae} averages the capped game scores to obtain RHAE. This is the scoring implementation used by the official scorecards \citep{arcagi3,arcscorecards2026}.

\section{AI Use Statement}
\label{app:ai-use}
We used generative AI tools to assist with creating and modifying scientific figures, writing and editing software code, drafting, polishing and proofreading the manuscript, and retrieving and identifying potentially relevant literature. We have not used generative AI tools to generate synthetic datasets, develop the theoretical or conceptual framework, formulate mathematical claims, contribute to or write proofs, propose or refine hypotheses, design the research methodology or experiments, perform translation, clean or reformat datasets, conduct qualitative or thematic analysis, or interpret the experimental results. LLM-generated code was verified and tested for correctness by the authors, all figures were checked against the underlying data, drafted text was edited and fact-checked by the authors, and LLM-identified literature was read and confirmed relevant before citation. We take responsibility for the final content of this work, including text, claims or artifacts produced with the aid of generative AI.

\end{document}